\documentclass[11pt]{article}
\usepackage{amsmath,amsthm,amssymb}
\usepackage{charter}
\usepackage{microtype}
\usepackage{graphicx,subfigure}
\usepackage{array,color} 
\usepackage{upgreek,bm} 
\usepackage{lettrine,algorithm,algpseudocode}

\def\0{\boldsymbol{0}}
\def\1{\boldsymbol{1}}
\def\x{\bm{x}}
\def\y{\bm{y}}
\def\n{\bm{n}}
\def\k{\bm{k}}

\def\s{\bm{s}}
\def\u{\bm{u}}
\def\a{\mathbf{a}}
\def\b{\mathbf{b}}
\def\c{\mathbf{c}}

\def\p{\mathbf{p}}
\def\e{\mathbf{e}}
\def\C{\mathbf{C}}
\def\R{\mathbf{R}}
\def\g{\mathbf{g}}
\def\K{\mathbf{K}}
\def\M{\mathbf{M}}
\def\A{\mathbf{A}}
\def\B{\mathbf{B}}
\def\I{\mathbf{I}}
\def\L{\mathbf{L}}
\def\J{\mathbf{J}}
\newcommand{\sinc}{\mathrm{sinc}}
\newcommand{\w}{\omega}
\newcommand{\abs}[1]{\left\lvert #1\right\rvert}
\newcommand{\norm}[1]{\left\lVert #1\right\rVert}
\newcommand{\h}{\widehat}
\newcommand{\bw}{\boldsymbol{\omega}}
\newcommand{\btau}{\boldsymbol{\tau}}
\newcommand{\bbeta}{\beta}
\DeclareMathOperator{\rect}{\mathrm{rect}}

\newtheorem{theorem}{Theorem}[section]

\newtheorem{proposition}[theorem]{Proposition}

\long\def\revu#1{{\color{black} #1}}
\long\def\Rev#1{{\color{black} #1}}

\begin{document}

\date{}

\title{Fast space-variant elliptical filtering \\ using box splines}

\author{Kunal Narayan Chaudhury\thanks{kchaudhu@math.princeton.edu}, Arrate Munoz-Barrutia, Michael Unser}

\maketitle

\begin{abstract}

   	The efficient realization of linear space-variant (non-convolution) filters is a challenging computational problem in image processing. In this paper, we demonstrate that it is possible to filter an image with a Gaussian-like elliptic window of varying size, elongation and orientation using a fixed number of computations per pixel.  The associated algorithm, which is based on a family of smooth compactly supported piecewise polynomials, the \textit{radially-uniform box splines}, is realized using pre-integration and local finite-differences. 
	
	The radially-uniform box splines are constructed through the repeated convolution of a fixed number of box distributions, which have been suitably scaled and distributed radially in an uniform fashion. The attractive features of these box splines are their asymptotic behavior, their simple covariance structure, and their quasi-separability. They converge to Gaussians with the increase of their order, and are used to approximate anisotropic Gaussians of varying covariance simply by controlling the scales of the constituent box distributions. Based on the second feature, we develop a technique for continuously controlling the size, elongation and orientation of these Gaussian-like functions. Finally, the quasi-separable structure, along with a certain scaling property of box distributions, is used to efficiently realize the associated space-variant elliptical filtering, which requires $O(1)$ computations per pixel irrespective of the shape and size of the filter.    
	
\end{abstract}

\textbf{Index} -- Space-variant filter,  Finite-differences, Running-sums, Anisotropic Gaussian, Box spline, Zwart-Powell (ZP) element.

\section{INTRODUCTION}
\label{I}

	The most widely used smoothing operator in image processing is the Gaussian filter. As far as  isotropic Gaussians  are concerned, a fast implementation is achievable simply by decomposing the filter into two orthogonal $1$-D Gaussians operating along the image axes. The $1$-D filters are in turn implemented using efficient recursive algorithms, e.g., the ones proposed by Deriche \cite{Deriche} and Young et al. \cite{Young}. \revu{We refer the readers to this survey article \cite{Tan2003} for an exhaustive account of such recursive schemes.} 
	
		A fundamental limitation of isotropic filtering is that it does not take into account the anisotropic nature of image features, which results in blurring of oriented patterns and textures. The development of fast anisotropic filtering in general, and  anisotropic Gaussian filtering in particular, have therefore gained momentum over the past decade. Worth mentioning in this regard is the work of Geusebroek et al. \cite{Smeulders}, who developed an efficient recursive technique based on the \Rev{factorization of the 2D-anisotropic Gaussian} into two $1$-D Gaussians, one along the image axes and the other along a generally off-grid direction. A drawback of this technique is that one has to interpolate the image along the off-grid direction to implement the corresponding $1$-D filter. To avoid interpolation and, in effect, to improve the spatial homogeneity and the Gaussian-like structure of the filters in \cite{Smeulders}, Lam et al. came up with the alternative ``triple-axis'' solution. Instead of using two directions, they chose to decompose the anisotropic Gaussian into three $1$-D Gaussians operating along one ofthe four cardinal directions: the horizontal, the vertical, and the two diagonals \cite{Lam2007}. The focus of these papers has largely been on space-invariant filtering, where the entire image is convolved with a single anisotropic Gaussian. \Rev{On the other hand,} a variety of space-variant filtering strategies have also been developed, including image statistics driven filtering \cite{Lee1980}, non-linear diffusion filtering \cite{PeronaMalik,Weikert1996} and gradient inverse-weighted filtering \cite{Wang1992}, to name a few.

\subsection{Linear space-variant filtering}

	 	In this paper, we focus on the paradigm of linear space-variant filtering using Gaussian-like kernels of different shapes and sizes. From a purely discrete\footnote{We associate the term ``discrete'' with functions defined on the Cartesian lattice $\mathbf{Z}^d$, where $d$ is the dimensionality of the function.} perspective, this calls for the design of a family of  Gaussian filters $\{g_{\lambda}[\n]\}_{\lambda}$, so that, given an input image $f[\n]$, one can evaluate the filtered samples $\bar{f}[\n]$ through the averaging
\begin{equation}
\label{discrete}
\bar{f}[\n]=\sum_{\k} f[\k] g_{\lambda(\n)} [\n-\k].
\end{equation}
The parameter $\lambda(\n)$, which specifies the covariance of the filter applied at location $\n$, allows one to continuously adjust the scale, orientation and elongation of the filter in keeping with the anisotropy of the local image features. There are however certain practical problems involved in an efficient realization of \eqref{discrete}. It is obvious that \eqref{discrete} cannot be written as a convolution, and hence cannot be realized using the FFT algorithm. In fact, the available options are either (i) to compute the filters $g_{\lambda}[\n]$ by sampling the continuous Gaussian on-the-fly, or (ii) to discretize $\lambda$ a priori, and to store the pre-compute filters in a look-up table. The problem with the former is that it proves to be extremely slow for wide kernels, while the latter restricts the control on the anisotropy of the filters. By appropriately modifying the recursive filtering strategies in \cite{Deriche, Young}, Tan et al. developed an algorithm for realizing \eqref{discrete} for the particular case where $\{g_{\lambda}[\n]\}_{\lambda}$ are isotropic \cite{Tan2003}.

	Spline kernels can also yield efficient algorithms for space-variant filtering, particularly when the space-variance is in terms of the scale (or size) of the kernel. For instance, Heckbert  proposed an algorithm for adaptive image blurring using tensor-products of polynomial splines, where the image is filtered with kernels of various scales using repeated integrations and finite-differences \cite{Heckbert}. \Rev{Based on similar principles, namely, the scaling properties of B-splines, Mu{\~{n}}oz-Barrutia et al. have developed an algorithm for fast computation of the continuous wavelet transform of $1$-D signals  \cite{Munoz}. Recently, the method was extended to perform space-variant filtering using Gaussian-like functions of arbitrary size, which can be elongated along the image axes \cite{Arrate_TIP}. To achieve this, the authors choose to approximate the Gaussian using separable B-splines. We propose to take this approach one step further. In particular, we overcome the limited steerability and ellipticity of the separable B-splines by considering certain quasi-separable analogues of B-splines, the so-called box splines \cite{deBoor}. As was demonstrated for the separable filters in \cite{Arrate_TIP}, we show that  these quasi-separable box splines can also be used to approximate the Gaussian, and that the associated space-variant filter can be decomposed into recursive pre-filters and scale-dependent finite difference filters. These together allow us to achieve a fast space-variant filtering of images using elliptical Gaussian-like filters.}

	To date, there have only been few applications of such multivariate splines in the fields of image processing and computer graphics. Noteworthy among them are the works of Richter \cite{Richter} and Asahi et al. \cite{Asahi}, concerning the development of image approximation and reconstruction algorithms, and that of Condat et al. \cite{Condat} and Entezari et al. \cite{Entezari2008}, concerning the development of interpolation formulas for hexagonal and BCC lattices.

\subsection{Main idea}

	In this contribution, we propose a fast space-variant filtering algorithm using a family of Gaussian-like box splines whose size, elongation and orientation can be continuously controlled. The attractive feature of our approach is that we use a continuous-discrete formalism which avoids the necessity of sampling a continuously-defined filter on-the-fly, or of storing a discrete set of pre-computed kernels. The developments in the paper are centered around two main ideas, which are as follows: \newline
	
\noindent (1) \textbf{The use of quasi-separable box splines}. The construction of bivariate box splines, conceived as the ``shadow'' of $N$-dimensional ($N \geq 2$) polytopes in $2$-dimensions, often turns out to be rather intricate (see \cite{deBoor} for instance). In this paper, we consider an alternative straightforward recipe for constructing box splines, namely, through repeated convolutions of dilated and rotated box distributions (see Fig. \ref{construction_zp}). In particular, we realize the so-called radially-uniform box spline $\bbeta^N_{\a} (\x)$ through the convolution of $N$ rotated box distributions, where $\a=(a_1,\ldots,a_N)$ is a scale-vector with $a_k$ being the scale of the box distribution positioned along the direction $(k-1)\pi/N$ (see \S\ref{III} for a precise definition). 

	The reason why the radially-uniform box splines are of interest in the current context is twofold. The first of these is that we can make them arbitrarily close to a Gaussian by increasing $N$ (see \S\ref{convergence}). The second reason, which has a more practical significance, is that we can continuously control their size, elongation, and orientation simply by acting on the scales. \newline

\begin{figure*}
\centering
\includegraphics[width=0.7\linewidth]{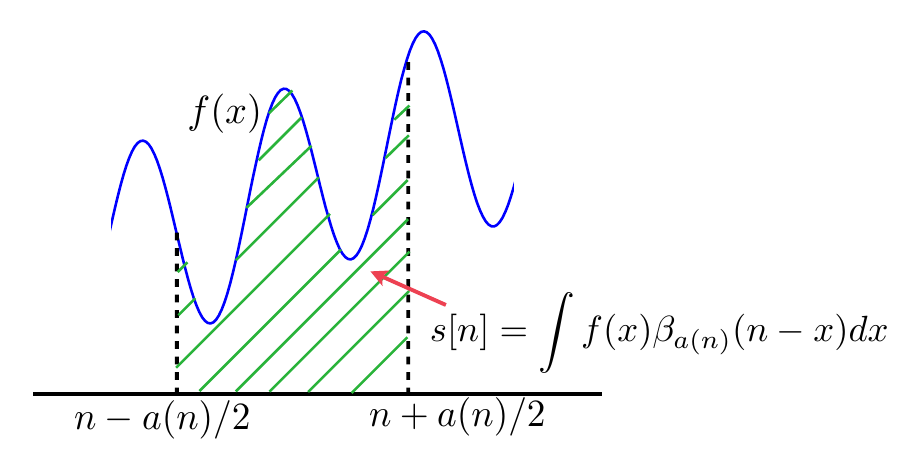}
\caption{Computation of the space-variant average $\bar{f}[n]$: The signal $f(x)$ is first localized (hatched zone) using the shifted box function $\beta_{a(n)}(n-x)$, and then the area  is computed. The central idea of our algorithm is to determine this area by taking the finite-difference of the primitive of $f(x)$.}
\label{main_idea}
\end{figure*}

\noindent (2) \textbf{An efficient strategy for space-variant averaging}. To convey this idea, we examine the following formula 
\begin{equation}
\bar{f}[n]=\frac{1}{2W(n)+1}\sum_{k=-W(n)}^{W(n)} f[n-k]
\label{simple_formula}
\end{equation}
for computing the space-variant averages of a discrete $1$-dimensional signal $f[n]$. We interpret the factor $W(n)$, which controls the amount of smoothing, as the size of the ``discrete box filter'' applied at location $n$. The disadvantage of using \eqref{simple_formula} is that its computational cost scales linearly with $W(n)$, which even gets worse in higher dimensions. This can be circumvented (with a mild interpolation cost) by considering instead the space-variant averaging
\begin{equation}
\label{SVF}
\bar{f}[n]=\frac{1}{a(n)} \int_{n-a(n)/2}^{n+a(n)/2} \! \! \! f(y)dy=\int f(y) \beta_{a(n)}(n-y)  dy,
\end{equation}
where we have replaced the discrete signal $f[n]$ by its interpolated version $f(x)$, and the discrete box filter by the box function 
\begin{equation*}
\beta_a(x)=
\begin{cases}  1/a &  \text{for \ $-a/2 <  x \leq  a/2$,} \\
0 &\text{otherwise}.
\end{cases}
\end{equation*}
The main advantage of this formulation is that we can realize \eqref{SVF} using $O(1)$ computations per position, independent of the size of $a(n)$. This is  based on the observation that \eqref{SVF} can computed by first evaluating the primitive 
\begin{equation*}
\displaystyle F(x)=\int_{-\infty}^{x} f(y) dy,
\end{equation*}
and then using the formula
\begin{equation}
\label{key_formula}
\bar{f}[n]=\frac{1}{a(n)} \Big(F(n+a(n)/2)-F\big(n-a(n)/2)\Big),
\end{equation}
which requires one addition and multiplication per position. This idea is illustrated in figure \ref{main_idea}. The other advantage is that, as opposed to the integer-valued window $W(n)$ in \eqref{simple_formula}, this gives access to the real-valued scale $a(n)$ for continuously controlling the smoothing. Indeed, if $f(x)$ is integrable (at least locally), then it can be shown that the use of small scales results in less smoothing, namely that $\bar{f}[n] \longrightarrow f[n]$ as $a(n) \longrightarrow 0$, whereas $\bar{f}[n]$ can be made negligibly small by making $a(n)$ sufficiently large.

	The contribution of this paper is the generalization of the filtering strategy in \eqref{SVF} to the bivariate setting using the radially-uniform box splines. In particular, given a discrete image $f[\n]$, we consider the space-variant filtering 
\begin{equation}
\label{SVF_2D}
 \bar{f}[\n]=\int f(\y)\bbeta_{\a(\n)}^N(\n-\y) d \y
\end{equation}
where $f(\x)$ represents a suitable interpolation of $f[\n]$. The significance of the quasi-separable characterization of $\bbeta_{\a}^N(\x)$ in terms of the box distributions is that it allows us to relate \eqref{SVF_2D} to the $1$-D problem in $\eqref{SVF}$. Indeed, we demonstrate in \S\ref{III} that \eqref{SVF_2D} can be implemented using an appropriate bivariate extension of pre-integrations and finite-differences, together with few evaluations of a fixed piecewise polynomial function (the coefficients of which are pre-computed). Although the derivation of the algorithm is rather involved, the final solution turns out to be remarkably simple (see \S\ref{V}, Algorithm  \ref{algo1}), and easy to implement.

\subsection{Notations}

We use  $\hat{f}(\bw)$ to denote the Fourier transform of a function $f(\x)$ on $\R^d$, specified by $\hat{f}(\bw)=\int_{\mathbf{R}^d} f(\x) \exp{(-j \bw^T \x)} d\x$. We use $f(\cdot-\boldsymbol{s})$ to denote the function
obtained by translating $f(\x)$ by $\boldsymbol{s}$. The convolution of two functions $f(\x)$ and $g(\x)$ is given by $(f \ast g)(\x)=\int_{\R^d} f(\s) g(\x-\s) d\s$. 
The notation $\circledast_{k=1}^N f_k (\x)$ is used to denote the convolution of a collection of functions $f_1(\x),\ldots,f_N(\x)$;  the order of the convolutions is immaterial. We suppress the domain of an integral (or summation) if this is obvious from the context. For a bivariate function $f(\x)=f(x_1,x_2)$, the partial derivative along $x_i$ is denoted by $\partial_i f(\x)$. \revu{Given operators $T_1$ and $T_2$ on a domain $\mathcal D$, we use $T_1 \circ T_2$ (often simply $T_1T_2$) to denote their composition: $(T_1 \circ T_2)(f)=T_1 (T_2(f))$ for every $f$ in $\mathcal D$. We use $\M^n$ to denote the $(n-1)$-fold matrix multiplication of $\M$ with itself. The integral $\int \M(\x) f(\x) d\x$, corresponding to a real-valued function $f(\x)$ and a matrix-valued function $\M(\x)$ on $\R^2$, denotes a matrix of the same dimension as $\M(\x)$, whose $(i,j)$-th component is given by $\int \M_{i,j}(\x) f(\x) d\x$. If $\mathbf P$ and $\mathbf Q$ are constant matrices, we then have $\int \mathbf{P} \M(\x)\mathbf{Q} f(\x)  d\x=\mathbf{P} (\int  \M(\x) f(\x) d\x) \mathbf{Q}$.} The  notation $f(x)=O(g(x)), x \longrightarrow 0$, signifies that there exists a constant $C$ (independent of $x$) such that $| f(x)| \leq  C g(x)$ for sufficiently small $x$.  The space of bivariate finite-energy signals is denoted by $\mathbf{L}^2(\R^2)$, or simply as $\mathbf{L}^2$; it is equipped with the norm $\lVert f \lVert_{\mathbf{L}^2}=[\int | f(\x)|^2 d\x] ^{1/2}$. The Dirac distribution is denoted by $\delta(\x)$.

\section{SPACE-VARIANT AVERAGING}
\label{II}

	We now derive \eqref{key_formula} \Rev{using an operator-based} formalism. This helps set up the framework needed for the subsequent generalization of the idea to higher dimensions and multiple orders in \S\ref{2Dalgo}.

\subsection{Realization of \eqref{SVF}}
\label{IIa}

\begin{figure*}
\centering
\includegraphics[width=1.0\linewidth]{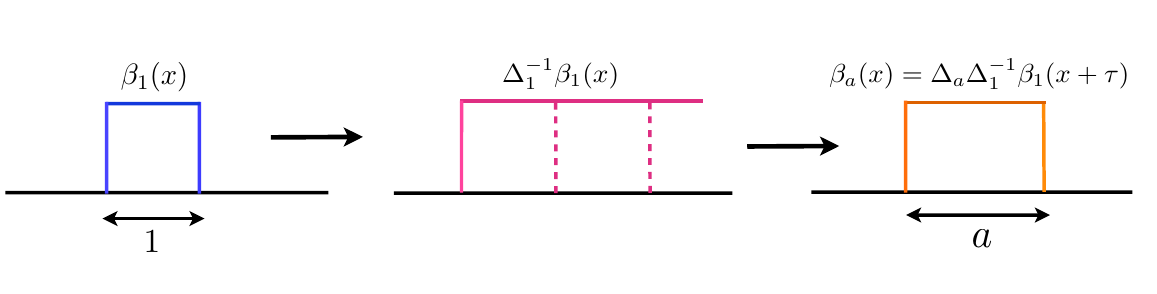}
\caption{Box function rescaling through ``addition and subtraction'' of the unit box function: The step function is first reproduced from the unit box using the running-sum, and then the appropriate finite-difference is applied to recover the rescaled box function.}
\label{box_function}
\end{figure*}

We consider the \textit{finite-difference} (FD) and the \textit{running-sum} (RS) of a function $f(x)$, which are respectively defined by
\begin{equation}
\label{FD} 
\Delta_{a} f(x)=\frac{1}{a}\big(f(x)-f(x-a)\big),
\end{equation}
and
\begin{equation}
\label{rs}
\Delta_b^{-1} f(x)=b\sum_{k=0}^{\infty} f(x-bk).
\end{equation}
The positive real numbers $a$ and $b$ are the scales of the operators\footnote{The notation $\Delta^{-1}_b$ is justified by the fact that $\Delta_a \Delta_b^{-1}$ acts as the identity operator when $a=b$.}. We note that the operators $\Delta_{a}$ and $\Delta_b^{-1}$, which takes $f(x)$ into $\Delta_{a} f(x)$ and  $\Delta_b^{-1} f(x)$ respectively, are linear and translation-invariant, and that when $b$ is an integer, $\Delta^{-1}_b$ can also be applied to a sequence $f[n]$ through the well-defined operation $\Delta^{-1}_b f[n]=b\sum_{k=0}^{\infty} f[n-bk]$. In particular, $g[n]=\Delta^{-1}_1 f[n]$ can be implemented efficiently using the simple recursion $g[n]= g[n-1]+f[n]$, under appropriate boundary conditions \cite{Munoz}.

	The significance of these operators is that we can relate the variable-size box functions in \eqref{SVF} to the unit-width box function using the transformation
\begin{equation}
\label{link} 
\beta_{a}(x)=\Delta_{a} \Delta^{-1}_1 \beta_{1}(x+\tau) 
\end{equation}
where $\tau=(a-1)/2$. In particular, this means that box functions of variable widths can be derived from a fixed box function through the successive applications of running-sums and finite-differences (see Fig. \ref{box_function}). To derive \eqref{link}, we note that $\Delta^{-1}_1\beta_{1}(x)=\sum_{k=0}^{\infty} \beta_1(x-k)=u(x+1/2)$, where the step function $u(x)$ equals $1$ for $x>0$ and zero otherwise. The desired result follows immediately:
\begin{equation*}
\Delta_{a} \Delta^{-1}_1\beta_{1}(x+\tau)= \frac{1}{a}\big(u(x+a/2)-u(x-a/2)\big)= \beta_a(x).
\end{equation*}

	We use \eqref{link} to derive the algorithm for computing \eqref{SVF} as follows: Fix an arbitrary position $n$ and the corresponding $a(n)$ in \eqref{SVF}, and consider the function
\begin{equation*}
s(x)=\int f(y) \beta_1(x-y)  dy.
\end{equation*}
We claim that $\bar{f}[n]  = \Delta_{a(n)} \Delta^{-1}_1 s(n+\tau)$. Indeed, following the linearity and translation-invariance of $\Delta_{a(n)} \Delta^{-1}_b$, and using \eqref{link}, we  can write
\begin{align*}
\Delta_{a(n)} \Delta^{-1}_1 s(x+\tau)&=\int f(y) [ \Delta_{a(n)} \Delta^{-1}_1 \beta_{1}(x+\tau-y)]  d y \\
&=\int f(y) \beta_{a(n)}(x-y)  d y,
\end{align*}
which establishes our claim. 

Now if the input signal is discrete, of the form $f(x)=\sum_{n \in \mathbf{Z}} f[n] \delta(x-n)$, then $s(x)$ can simply be written as $s(x)=\sum f[n] \beta_1(x-n)$. A simple manipulation then shows that $\Delta^{-1}_1 s(x)=\sum g[n] \beta_1(x-n)$, where $g[n]$ is the running-sum of $f[n]$. Thus, denoting the piecewise-constant interpolation of $g[n]$ by $F(x)$, we obtain
\begin{align*}
\bar{f}[n]=\Delta_{a(n)} F(n+\tau).
\end{align*}
This leads us to the following two-step algorithm for realizing  \eqref{SVF}:

\begin{description}
\item (1) (\textbf{Space-invariant step}) Compute $g[n]=\Delta^{-1}_1 f[n]$ using the recursion $g[n]=g[n-1]+ f[n]$;
\item (2) (\textbf{Space-variant step}) For every position $n$, set $\bar{f}[n]=\Delta_{a(n)} F(n+\tau)$, where $\tau=(a(n)-1)/2$.
\end{description}

\revu{The steps of the algorithm can be visualized for the particular case when the input is an impulse and when $a(n)=a$ for every $n$ using Fig. \ref{box_function}. The second and third plots in this figure then correspond to steps (1) and (2) of the algorithm, respectively.} 

The remarkable fact about the algorithm is that the space-variant aspect of the transformation $f[n] \mapsto \bar{f}[n]$ gets transferred to the scale-dependent operator $\Delta_a$, which in turn is implemented at a fixed computational cost per pixel, namely, one addition and multiplication per position. We would also like to note that \eqref{link} can more generally be expressed as 
\begin{equation}
\label{link_general} 
\beta_{a}(x)=\Delta_{a} \Delta^{-1}_b \beta_{b}(x+\tau)    \qquad (\tau=(a-b)/2).
\end{equation}
The significance of this relation is that, if the lattice spacing $b$ is different from unity, one can still realize the running-sum (without interpolation) by replacing the operator $\Delta^{-1}_1$ by $\Delta^{-1}_b$. We will use this in the sequel.

\subsection{Bivariate extension}
\label{III}

\subsubsection{Radially-uniform box splines}

\begin{figure*}
\centering
\includegraphics[width=0.7\linewidth]{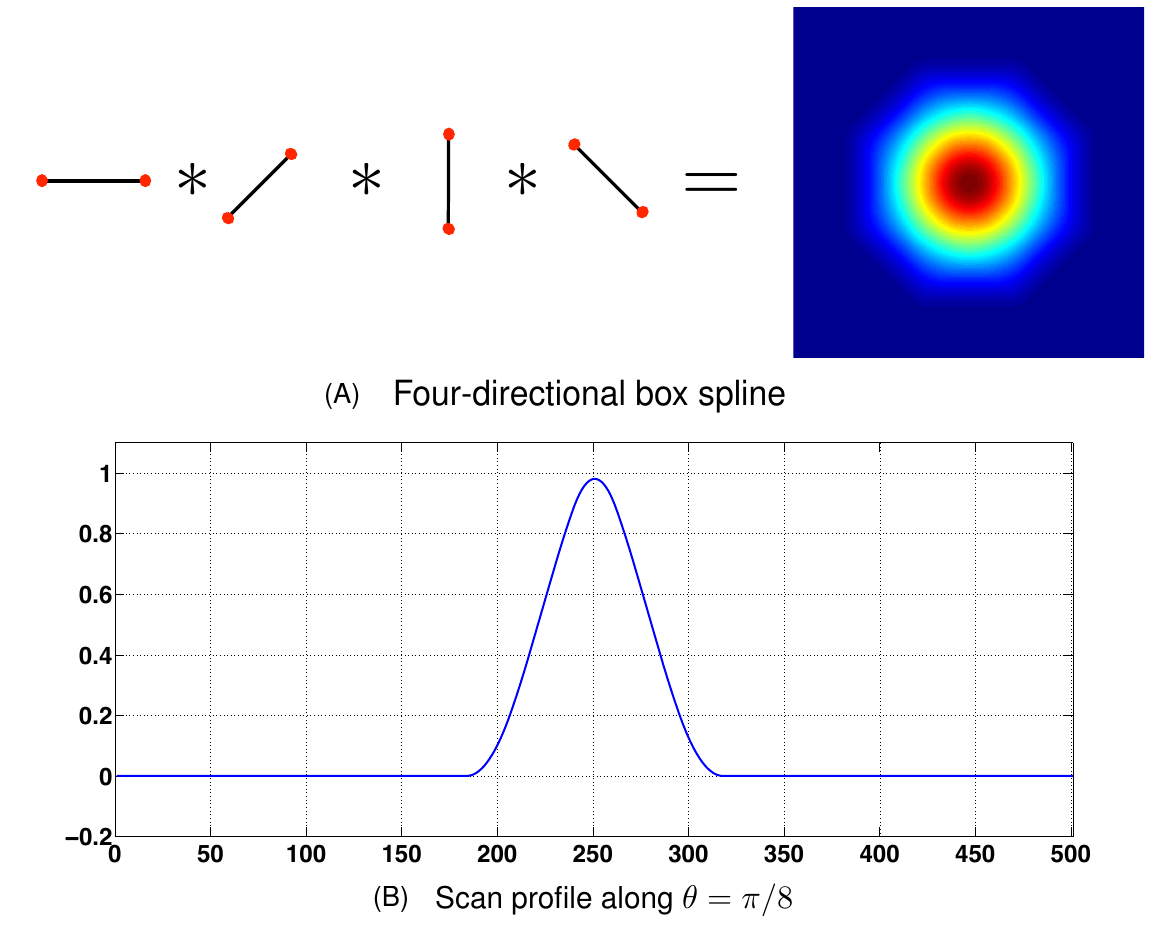}
\caption{Construction of the radially-uniform box spline through the convolution of four directional box distributions. (\textbf{A}) The four box distributions, distributed uniformly over $[0,\pi)$, were assigned equal scales in this example; (\textbf{B}) Scan profile along $\theta=\pi/8$.}
\label{construction_zp}
\end{figure*}

	We now extend the space-variant filtering strategy discussed in the previous section to the bivariate setting, where the additional notion of directionality is appropriately addressed. As a first step, we devise an appropriate directional extension of the box function. In particular, corresponding to a real-valued scale $a$ and direction $\theta$, we define the \textit{directional box distribution} $\varphi_{a,\theta}(\x)$ as the tensor product of the box function $\beta_a(x)$ and the Dirac distribution $\delta(x)$ operating along orthogonal directions,
\begin{equation*}
\varphi_{a,\theta}(\x)=\beta_a \left(\u_{\theta}^T\x\right)\delta \left(\u_{\theta_{\perp}}^T \x \right).
\end{equation*} 
Here the orthogonal directions are specified by the unit vectors $\u_{\theta}=(\cos\theta, \sin\theta)$, and $\u_{\theta_{\perp}}=(-\sin\theta, \cos\theta)$. The scale $a$ controls the amount of smoothing applied along the orientation of the box distribution, whereas no smoothing is applied along the transverse direction. The idea then is to construct the box spline by convolving an arbitrary number of such directional box distributions (cf. Fig \ref{construction_zp}). Thus, corresponding to an integer $N > 1$, a set of real-valued scales $a_1,\ldots,a_N$, and uniform rotation-angles $\theta_k=(k-1)\pi/N, k=1, \ldots,N$, we specify the \textit{radially-uniform box spline} through the convolution
\begin{equation}
\label{const}
\bbeta^N_{\a} (\x)= (\varphi_{a_1,\theta_1}\ast \cdots \ast \varphi_{a_N,\theta_N})(\x).
\end{equation}
We shall refer to $N$ and the tuple $\a=(a_1,\ldots,a_N)$ as the directional-order and the scale-vector of the box spline, respectively.

	Following definition \eqref{const}, it can be verified that $\bbeta^N_{\a}(\x)$ is a piecewise polynomial of degree $\leq N-2$, where the partitions are specified by lines running along the directions $\theta_1,\ldots,\theta_N$. Moreover, $\bbeta^N_{\a}(\x)$ is symmetric with respect to the origin, and is compactly supported on a convex $N$-sided polygon consisting of the points
\begin{equation*}
\Big\{\sum_{k=1}^N t_k a_k \u_{\theta_k}:  \ -1/2 \leq t_k \leq1/2 \Big\}.
\end{equation*}
The radially-uniform box splines are non-separable for $N > 2$. However, in keeping with the spirit of the underlying tensor construction, the term quasi-separable would be more appropriate. The scale-vector $\a$ plays a vital role in determining  the size and shape of  the box spline. It is clear that the box spline can be arbitrarily elongated along the principal directions $\theta_k$ $(1 \leq  k  \leq  N)$ simply by rescaling the box distribution $\varphi_{a_k,\theta_k}$, that is by making $a_k$ large compared to the other scales. Moreover, we will demonstrate in the sequel that one can elongate the box spline along \textit{any} arbitrary direction by appropriately acting on the scale-vector. The role of the directional-order is more subtle; it determines the degrees of freedom available for controlling the geometry of the box spline and also its regularity (smoothness). We will discuss these aspects in detail for the particular four-directional box spline ($N=4$) in \S\ref{convergence}.

\subsubsection{Realization of \eqref{SVF_2D}}
\label{2Dalgo}
	
	We now formulate the algorithm for realizing \eqref{SVF_2D} by appropriately extending the domain of definition of the FD and the RS operator to bivariate functions. The main idea is to derive a relation similar to \eqref{link} for the radially-uniform box splines. Thus, corresponding to positive real-valued scales $a$ and $ b$, and  direction $0 \leq  \theta <\pi$, we consider the \textit{directional finite-difference} 
\begin{equation}
\label{FD1}
\Delta_{a,\theta}f (\x)=\frac{1}{a}\Big(f(\x)-f(\x-a\u_{\theta})\Big),
\end{equation}
and the \textit{directional running-sum}
\begin{equation}
\label{RS1}
\Delta^{-1}_{b,\theta}f(\x)=b\sum_{k=0}^{\infty} f(\x- k b \u_{\theta}).
\end{equation}
In keeping with the definition of the box spline, the radially-uniform FD and RS operators, $\Delta^N_{\a}$ and $\Delta^{-N}_{\b}$, are then specified by the combined action of \eqref{FD1} and \eqref{RS1} along the directions $\theta_k=(k-1)\pi/N$. In particular, we set
\begin{equation}
\label{2DFD}
\Delta^N_{\a}=\Delta_{a_1,\theta_1}  \circ \cdots \circ \Delta_{a_N,\theta_N},
\end{equation}
and
\begin{equation}
\label{2DRS}
 \Delta^{-N}_{\b}=\Delta^{-1}_{b_1,\theta_1} \circ \cdots \circ \Delta^{-1}_{b_N,\theta_N},
\end{equation}
where the scale-vectors $\a=(a_1,\ldots,a_N)$ and $\b=(b_1,\ldots,b_N)$ specify the scale along each direction. The operators $\Delta^N_{\a}$ and $\Delta^{-N}_{\b}$ are closely related to the radially-uniform box splines. Indeed, it can readily be verifed that
\begin{equation*}
\Delta^N_{\a}   \Delta^{-N}_{\b}
=\Delta_{a_1,\theta_1}\Delta^{-1}_{b_1,\theta_1}  \circ \cdots  \circ \Delta_{a_N,\theta_N}  \Delta^{-1}_{b_N,\theta_N},
\end{equation*}
and that using \eqref{link_general} we can write
\begin{equation*}
\Delta_{a,\theta} \Delta^{-1}_{b,\theta} \varphi_{b,\theta} (\x+\tau \u_{\theta} )=  \varphi_{a,\theta} (\x).
\end{equation*}
Thus, if we let $\btau=\sum  \tau_k \u_{\theta_k}$, where $\tau_k=(a_k-b_k)/2$, then following definitions \eqref{const}, \eqref{2DFD}, and \eqref{2DRS}, we see that
\begin{align*}
&\Delta^N_{\a} \Delta^{-N}_{\b} \bbeta^N_{\b}(\x+\btau) \\
&=\Delta_{a_1,\theta_1}\Delta^{-1}_{b_1,\theta_1}  \circ \cdots \circ \Delta_{a_N,\theta_N}  \Delta^{-1}_{b_N,\theta_N} \Big[ \circledast_{k=1}^N \varphi_{b_k,\theta_k} (\x+\tau_k \u_{\theta_k}) \Big] \\
&=\circledast_{k=1}^N \Delta_{a_k,\theta_k}\Delta^{-1}_{b_k,\theta_k} \varphi_{b_k,\theta_k} (\x+\tau_k \u_{\theta_k})  \\
&=\displaystyle \circledast_{k=1}^N  \varphi_{a_k,\theta_k} (\x).
\end{align*}
This provides the following crucial connection between the box splines and the $2$-D operators.
\begin{proposition}
\textit{The box spline $\bbeta^N_{\a}(\x)$ can be expressed as
\begin{equation}
\label{link2}
\bbeta^N_{\a}(\x) =\Delta^N_{\a}  \Delta^{-N}_{\b} \bbeta^N_{\b}(\x+\btau).
\end{equation}}
\end{proposition}
Before discussing the filtering algorithm, we briefly elaborate on the implementation of $\Delta^N_{\a}$ and $\Delta^{-N}_{\b}$. We can show that \eqref{2DFD} can be  written as
\begin{equation}
\label{FD_mesh}
\Delta_{\a}^{N}f(\x)=\sum_{i=0}^{2^{N}-1} w_i f(\x-\x_{i}),
\end{equation}
where $w_i= (-1)^{q_1+\cdots+q_N} (a_1\cdots a_N)^{-1}$ and $\x_i=\sum_{k=1}^N q_k a_k \u_{\theta_{k}}$, and the index $i$ taking values between $0$ and $(2^N-1)$ is the decimal counterpart of the binary number $(q_N,q_{N-1},\ldots,q_1)$, which takes values from $(0,\ldots,0)$ to $(1,\ldots,1)$. We identify the points $\x_i$ with the vertices of an affine mesh, and $w_i$ with the corresponding mesh taps (cf. Fig. \ref{mask}). 

   	As far as the application of $\Delta^{-N}_{\b}$ to a discrete sequence $f[\n]$ is concerned, the unfortunate difficulty is that the vectors $b_k \u_{\theta}$ must necessarily lie on the lattice for \eqref{RS1} to be well-defined. In fact, it is easily seen that one cannot associate a digital filter with the RS operators in general. However, the good news is that, when $N$ equals $2$ and $4$, the transformation $f[\n] \mapsto \Delta^{-N}_{\b} f[\n]$ can be exactly realized without the need for interpolation by appropriately setting the scales $b_k$ of the directional running-sums. We will discuss the latter case in detail in \S\ref{V}.

	The algorithm for realizing \eqref{SVF_2D} corresponding to a specified scale-vector map $\a(\n)$ is based on relation \eqref{link2}. In particular, by considering the function $f(\x)=\sum_{\n} f[\n] \delta(\x-\n)$, and by proceeding exactly along lines of the $1$-D derivation, we express the filtered samples in \eqref{SVF_2D} as 
\begin{equation} 
\label{keyeqn}
\bar{f}[\n]=\sum_{i=0}^{2^N-1} w_i F(\n+\btau-\x_i),
\end{equation} 
 where $F(\x)=\sum g_{\b}[\n] \bbeta^N_{\b}(\x-\n)$ denotes the interpolated version of the pre-integrated signal $g_{\b}[\n]=\Delta^{-N}_{\b} f[\n]$; $\btau=0.5(\sum (a_k(\n)-b_k)\cos \theta_k, \sum (a_k(\n)-b_k)\sin \theta_k)$; and the pairs $(\x_i, w_i)$ are the vertices and taps of the affine FD mesh in \eqref{FD_mesh}. Note that $\btau, w_i$ and $\x_i$ are defined pointwise in \eqref{keyeqn}; we dropped the index $\n$ to simplify the notation. We will discuss the implementation aspects of the algorithm, particularly the computation of $g_{\b}[\n]$ and its interpolated form $F(\x)$ for the case $N=4$ in \S\ref{V}.

\subsubsection{Characterization of the radially-uniform box splines}
\label{convergence}

	 The motivation behind introducing the radially-uniform box splines was to develop elliptical Gaussian-like filters, whose shape (size, elongation and orientation) can be continuously controlled, and a fast space-variant algorithm using such filters. Indeed, it turns out that the radially-uniform box splines (and its iterated versions) form close approximates of the Gaussian. To substantiate our claim we present the following result (proof in Appendix \S\ref{appendix A}) that can well be seen as a ``radial'' version of the central limit theorem. 	   
	  
\begin{theorem} 
\label{convergenceOrder}
\textit{Consider the sequence of box splines $\bbeta^2_{\a(2)}(\x),\bbeta^3_{\a(3)}(\x),\ldots$ corresponding to the scale-vectors $\a(2), \a(3), \ldots$, where $a_k(N)= \sigma \surd(24/N)$ for $1 \leq k \leq N$. Then the following holds 
\begin{equation}
\label{conv1}
 \lim_{N \longrightarrow \infty} \bbeta^N_{\a(N)}(\x)=\frac{1}{2\pi \sigma^2} \exp\Big(-\frac{\norm{ \x}^2}{2\sigma^2} \Big).
\end{equation}}
\end{theorem}
In fact, the radially-uniform box splines constructed using uniform scale-vectors are supported on a $N$-sided uniform polygon, and it can be shown that they have  continuous derivatives of order $(N-3)$. The above result is then consistent with the fact that the isotropy and smoothness of such box splines progressively improves with the increase in the directional-order. Moreover, it is also possible to mimic certain anisotropic Gaussians by using a sequence of non-uniform scale-vectors. Indeed, as a direct extension of Theorem \ref{convergenceOrder}, one can construct sequences of box splines which converge to anisotropic Gaussians as $N$ increases. 
	
	Yet another useful form of anisotropic convergence is achievable based on the serial convolutions of a radially-uniform box spline, of a fixed directional-order, with itself. In particular, corresponding to fixed integers $N$ and $m$ ($m \geq 1$), and a scale-vector $\a=(a_1,\ldots,a_N)$, we consider the \textit{iterated radially-uniform box spline} 
\begin{equation}
\bbeta^{N,m}_{\a}(\x)=(\bbeta^N_{\a} \ast \cdots \ast \bbeta^N_{\a})(\x)
\label{iterated_spline}
\end{equation}
obtained through the $(m-1)$-fold convolution of $\bbeta^N_{\a}(\x)$ with itself. Then, for the particular sequence of box splines $\{\bbeta^{N,m}_{\a(m)}(\x)\}$ corresponding to the scale-vectors $\a(m)=(a_1/\sqrt m,\ldots,a_N/\sqrt m)$, we have the result
\begin{equation}
\label{conv2}
\lim_{m \longrightarrow \infty} \bbeta^{N,m}_{\a(m)}(\x)= \frac{1}{2\pi \abs{\det(\C)}^{1/2}}\exp\Big(-\frac{1}{2} \x^T\C^{-1}\x\Big),
\end{equation}
where 
\begin{equation}
\label{cov}
\C=\frac{1}{12} \left(\begin{array}{cc} \sum a_k^2 \cos^2\theta_k & \frac{1}{2}\sum a_k^2 \sin2\theta_k  \\ \\  \frac{1}{2}\sum a_k^2 \sin2\theta_k &  \sum a_k^2 \sin^2\theta_k \end{array}\right).
\end{equation}
Indeed, this follows directly from a certain version of the central limit theorem, which tells us that the each of the components
\begin{equation*}
\circledast_{j=1}^m \ \varphi_{a_k/ \sqrt{m} ,\theta_k}(\x)  \qquad (1 \leq k \leq  N)
\end{equation*}
converge to a ``directional'' Gaussian distribution as $m \longrightarrow \infty$. The covariance $\C$ is then given by the limiting sum of the covariances $\C_k$ of the constituent box distributions. The utility of such iterated box splines will be discussed in \S \ref{discussion}.

\begin{figure}
\centering
\includegraphics[width=0.7\linewidth]{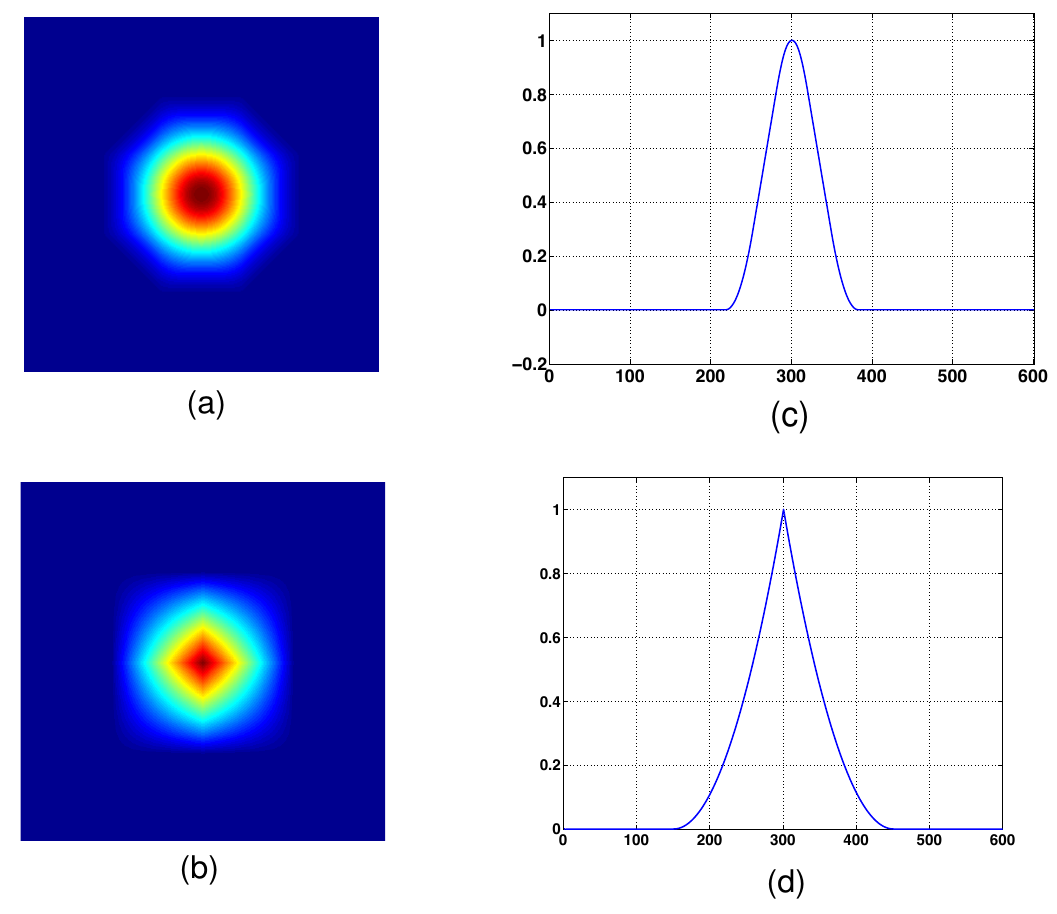}
\caption{Intensity distribution of  (a) the radially-uniform box spline, and (b) the separable B-spline, of order four. The respective scan profiles along $\pi/8$ are shown in (c) and (d). }
\label{construction_box_spline}
\end{figure}
			
	 Having characterized the asymptotic behavior of the box splines, we now focus on the problem of approximating an anisotropic Gaussian using a box spline of a fixed directional-order. Since a centered Gaussian is uniquely specified by its covariance, we propose a finite-order box spline approximation of the same based on its covariance. This, in fact, amounts to a moment-based approximation of the Gaussian, that is, the box spline resembles the Gaussian up to its second-order moments. Moreover, since the level-sets of Gaussians are ellipses, this equivalently amounts to constructing elliptical filters of different size, elongation, and orientation. 
	 
	 The covariance of the radially-uniform box spline, namely,
\begin{equation*}
\C^N_{\a}=\int \x \x^T \bbeta^N_{\a}(\x) d \x,
\end{equation*}
 can be expressed as the sum of the covariances of the box distributions (cf. Appendix \S \ref{appendix B} for details) as follows:
\begin{equation}
\label{covariance}
\C^N_{\a}=\frac{1}{12} \left(\begin{array}{cc} \sum a_k^2 \cos^2\theta_k &
\frac{1}{2}\sum a_k^2 \sin2\theta_k  \\ \\
\frac{1}{2}\sum a_k^2 \sin2\theta_k & \sum a_k^2 \sin^2\theta_k\end{array}\right).
\end{equation}  
This provides the explicit dependence of $\C^N_{\a}$ on the scale-vector. In particular, based on the eigen decomposition of $\C^N_{\a}$, we propose the following characterization of the elliptical parameters of the box spline: Let $\lambda_{\max}$ and $\lambda_{\min}$ denote the largest and smallest eigenvalues of $\C^N_{\a}$, and $(v_1,v_2)$ the eigenvector corresponding to the eigenvalue $\lambda_{\max}$. The size $s^N_{\a}$, elongation $\varrho^N_{\a}$, and orientation $\theta^N_{\a}$ of the radially-uniform box spline $\bbeta^N_{\a}(\x)$ are then defined as
\begin{align}
\label{def_eigen}
s^N_{\a}&=\lambda_{\max}+\lambda_{\min}=\frac{1}{12} \sum a^2_k,  \nonumber \\
\varrho^N_{\a}&=\frac{\lambda_{\max}}{\lambda_{\min}}=\frac{\sum  a^2_k+ \sqrt{D}}{\sum   a^2_k-\sqrt{ D}},  \nonumber \\
\theta^N_{\a}&=\tan^{-1}\left(\frac{v_2}{v_1}\right)=\tan^{-1} \left(\frac{-\sum   a^2_k \cos(2\theta_k)+\sqrt{D}}{\sum   a^2_k \sin(2\theta_k)}\right),
\end{align}
where $D=\left(\sum a_{k}^{2}\cos2\theta_{k}\right)^2+\left(\sum a_{k}^{2}\sin2\theta_{k}\right)^2$.

	Since $\C^N_{\a}$ is strictly positive (see Appendix \S\ref{appendix B}), all the above parameters are indeed well-defined. Note that the covariance matrix in \eqref{covariance} and the triple in \eqref{def_eigen} provide equivalent descriptions of the box spline geometry. The motivation behind introducing the latter is its convenient rotation-invariant nature: while $\C^N_{\a}$ changes with the rotations of a given box spline, $s^N_{\a}$ and $\varrho^N_{\a}$ remains fixed. It is for this reason that we use the latter description for studying the dependence of the elliptical geometry on the scale-vector in the next section.

\section{FOUR-DIRECTIONAL BOX SPLINES}
\label{V}

	We now study the particular \textit{four-directional} box spline
\begin{equation*}
\bbeta^4_{\a}(\x)=(\varphi_{a_1,0} \ast \varphi_{a_2,\pi/4} \ast \varphi_{a_3,\pi/2} \ast \varphi_{a_4,3\pi/4})(\x),
\end{equation*}
and the corresponding implementation aspects. This particular box spline is composed of patches of quadratic polynomials (degree $\leq$ 2), is continuously differentiable, and is compactly supported on a convex octagon (cf. Fig. \ref{construction_zp}).  

	\Rev{We note that in \cite{Arrate_TIP} the authors have used separable B-splines to approximate the Gaussian. Although these functions are built from the same constituent box distributions, the advantage of the four-directional box spline over the separable ones is that they are more isotropic. As seen in Fig. \ref{construction_box_spline}, the basic four-directional box spline, besides having a smaller support, exhibits a more Gaussian-like profile than the separable counterpart of identical order. In addition, the anisotropic four-directional box spline can be rotated to arbitrary orientations, while the separable ones are constrained to the image axes.}

\subsection{Fast space-variant elliptical filtering}
\label{filtering}

\begin{figure}
\centering
\includegraphics[width=0.7\linewidth]{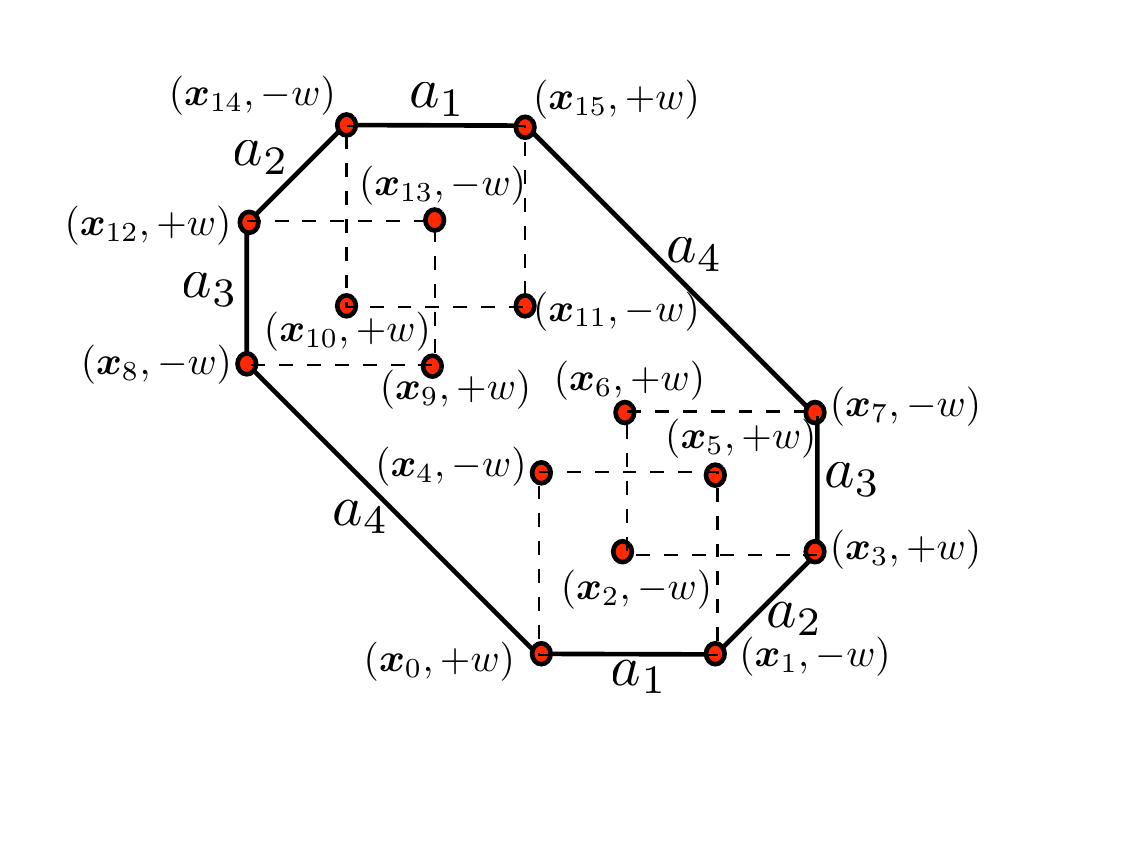}
\caption{The distribution of the taps of the FD mesh. The pairs $(u,v)$ denote the positions $(u)$ and the corresponding weights $(v)$ of the taps of the FD mesh.}
\label{mask}
\end{figure}

	The four-directional box spline is of particular interest in the context of the space-variant filtering following the fact that $\Delta^{-4}_{\b}$ can be implemented without interpolation when $\b=(1,\sqrt 2,1,\sqrt 2)$. The corresponding interpolating function, $\bbeta^4_{\b}(\x)$, turns out to be well-known in the box spline community, and is popularly referred to as the Zwart-Powell (ZP) element \cite{deBoor,Zwart}. The steps for realizing \eqref{SVF_2D} using the four-directional box spline are as follows: \newline
	
\noindent (1)  (\textbf{Pre-integration}) The crucial point is the choice of the scale-vector $\b=(1,\sqrt 2,1,\sqrt 2)$ corresponding to which the RS operator
\begin{equation*}
\Delta^{-4}_{\b}=\Delta^{-1}_{1,0} \circ \Delta^{-1}_{\sqrt 2,\pi/4} \circ \Delta^{-1}_{1,\pi/2} \circ \Delta^{-1}_{\sqrt 2,3\pi/4}
\end{equation*}
can be directly applied to $f[\n]$. In particular, the running-sum $g_{\b}[\n]=\Delta^{-4}_{\b} f [\n]$ can be computed using the following four steps.
\begin{description}
\item (RS1)  Horizontal running-sum: 
\begin{equation*}
g_0[n_1,n_2]=\Delta^{-1}_{1,0} f [n_1,n_2]=\sum_{k=0}^{\infty} f[n_1-k,n_2] .
\end{equation*}

\item (RS2)   First-diagonal running-sum: 
\begin{equation*}
g_{\pi/4}[n_1,n_2]=\Delta^{-1}_{\sqrt 2,\pi/4} g_0 [n_1,n_2]= \sqrt 2 \sum_{k=0}^{\infty} g_0[n_1-k,n_2-k] .
\end{equation*}

\item (RS3)   Vertical running-sum: 
\begin{equation*}
g_{\pi/2}[n_1,n_2]=\Delta^{-1}_{1,\pi/2} g_{\pi/4} [n_1,n_2]=\sum_{k=0}^{\infty} g_{\pi/4}[n_1,n_2-k].
\end{equation*}

\item (RS4)   Second-diagonal running-sum:
\begin{equation*}
g_{\b}[n_1,n_2]=\Delta^{-1}_{\sqrt 2,3\pi/4} g_{\pi/2} [n_1,n_2]=\sqrt 2 \sum_{k=0}^{\infty} g_{\pi/2}[n_1+k,n_2-k] .
\end{equation*}

\end{description}

\noindent (2) (\textbf{Finite-difference}) At each position $\n$, the FD mesh is computed using the scale-vector $\a(\n)$. The weights $w_i$ and the vertices $\x_i$ are listed in Table \ref{mask_positions} with the convention that $a'_k=a_k/\sqrt 2$ for $k=2,4$. The mesh has a total of $4\times 4=16$ vertices; in particular, there are $4$ clusters corresponding to the four boxes with $4$ vertices per cluster, as shown in Fig. \ref{mask}. The shift $\btau=(\tau_1,\tau_2)$ is given by $\tau_1=(\sqrt{2}a_1+a_2-a_4-\sqrt{2})/2\sqrt{2}$ and $\tau_2=(a_2+\sqrt{2}a_3+a_4-3\sqrt{2})/2\sqrt{2}$. The filtered sample is then computed using the formula
\begin{equation}
\label{final_eqn}
 \bar{f}[\n] = \sum_{i=0}^{15} w_i F(\n+\btau-\x_i).
\end{equation}
The interpolation samples $F(\x)=\sum g_{\b}[\n] \bbeta^4_{\b}(\x-\n)$ in \eqref{final_eqn} are computed efficiently by taking advantage of the piecewise polynomial structure of the compactly supported ZP element (see Appendix \ref{fast_zp}).

	As in the $1$-D setting, the running-sums are efficiently evaluated using recursions as summarized in Algorithm \ref{algo1}. The decisive computational advantage, especially for wider kernels, is derived from the fact that the number of vertices of the FD mesh is completely independent of the scale-vector. As a result, the algorithm has a fixed computational cost per pixel, modulo the cost of the running-sum and the interpolations (see Table \ref{computation_time}).

\begin{table}[!t]
\caption{\revu{Specification of the taps of the FD mesh associated with the operator $\Delta^4_{\a}$. The weight $w$ is given by $(a_1a_2a_3 a_4)^{-1}$, where $\a=(a_1,a_2,a_3,a_4)$ is the corresponding scale-vector.}} 
\label{mask_positions} \vspace{0.2cm}
\centering
\begin{tabular}{|c|c|c|c|c|c|}
\hline
\bfseries $i$ & \bfseries $\x_i$ & \bfseries $w_i$  & $i$ & \bfseries $\x_i$ & $w_i$ \\
\hline 
$0$ & $(0,0)$ &  $+w$ & $8$ & $(-a'_4,a'_4)$ & $-w$ \\
$1$ & $(a_1,0)$ & $-w$ & $9$ & $(a_1-a'_4,a'_4)$ & $+w$\\
$2$ & $(a'_2,a'_2)$ &  $-w$ & $10$ & $(a'_2-a'_4,a'_2+a'_4)$ & $+w$ \\
$3$ & $(a_1+a'_2,a'_2)$ & $+w$ &  $11$ & $(a_1+a'_2-a'_4,a'_2+a'_4)$ & $-w$\\
$4$ & $(0,a_3)$ & $-w$  & $12$ & $(-a'_4,a_3+a'_4)$ & $+w$ \\
$5$ & $(a_1,a_3)$ & $+w$  & $13$ & $(a_1-a'_4,a_3+a'_4)$ & $-w$ \\
$6$ & $(a'_2, a_3+a'_2)$ & $+w$ & $14$ & $(a'_2-a'_4,a_3+a'_2+a'_4)$ & $-w$ \\
$7$ & $(a_1+a'_2,a_3+a'_2)$ & $-w$ & $15$ & $(a_1+a'_2-a'_4,a_3+a'_2+a'_4)$ & $+w$ \\
\hline
\end{tabular}
\end{table}

\begin{algorithm}{}
\caption{Space-variant elliptical filtering}
\label{algo1}
\begin{algorithmic}
     \State 1. Input: $f[\n]$ and $\a(\n)$
     \State 2. Perform recursions:   \begin{description}
	\item $g_0[n_1,n_2] \leftarrow f[n_1,n_2]+g_0[n_1-1,n_2]$
	\item $g_{\pi/4}[n_1,n_2] \leftarrow \sqrt{2} g_0[n_1,n_2]+g_{\pi/4}[n_1-1,n_2-1]$
	\item $g_{\pi/2}[n_1,n_2] \leftarrow g_{\pi/4}[n_1,n_2]+g_{\pi/2}[n_1,n_2-1]$
	\item $g_{\b}[n_1,n_2] \leftarrow \sqrt{2} g_{\pi/2}[n_1,n_2]+g_{\b}[n_1+1,n_2-1]$
\end{description}
     \State 3. Local FD operation:
     \For{each position $\n$}
       \State  compute $w_i,\x_i$ and $\btau$ using $\a(\n)$
       \State evaluate  $F(\n+\btau-\x_i)$ using ZP interpolation
       \State $\bar{f}[\n] \leftarrow \sum_i w_i F(\n+\btau-\x_i)$
    \EndFor
     \State 4. Return $\bar{f}[\n]$
\end{algorithmic}
\end{algorithm}	

\subsection{Size, elongation and orientation of the box splines}
\label{geometry_control}

As was mentioned earlier, the size and shape of the radially-uniform box spline can be controlled by appropriately adjusting the scales of the constituent box distributions. In this regard, we now discuss the following: (i) the forward problem of controlling the anisotropy of the four directional box spline by acting on the scale-vector, and (ii) the inverse problem of uniquely specifying the scale-vector of the box spline corresponding to a given  covariance (geometry). For notational ease, we shall henceforth drop the superscript $N=4$ when referring to the four-directional box spline and its related parameters. 

\subsubsection{Control on the anisotropy}
         
          The elliptical geometry of this box spline is specified using parameters defined in \eqref{def_eigen}, namely,
\begin{equation*}
s_{\a}=\frac{1}{12}\sum a_k^2, \quad \theta_{\a}=\tan^{-1} \left(\frac{a^2_3-a^2_1+\sqrt{D}}{a^2_2-a^2_4}\right), \  \text{ and }  \ \varrho_{\a}= \frac{\sum a^2_k+\sqrt{D}}{\sum a^2_k-\sqrt{D}},
\end{equation*}
where $D=(a^2_3-a^2_1)^2+(a^2_2-a^2_4)^2$. It turns out that the size and orientation can be arbitrarily controlled by adjusting the scale-vector. Indeed, the size can be easily manipulated by multiplying the scale-vector $\a$ with a constant factor, since this leaves both the orientation and elongation unchanged. The elongation can be arbitrarily controlled in the neighborhood of the four principal directions. However, there exists a finite upper bound on the elongation along other directions (cf. Appendix \S\ref{proof:ellipticity}).

\begin{figure}
\centering
\includegraphics[width=0.6\linewidth]{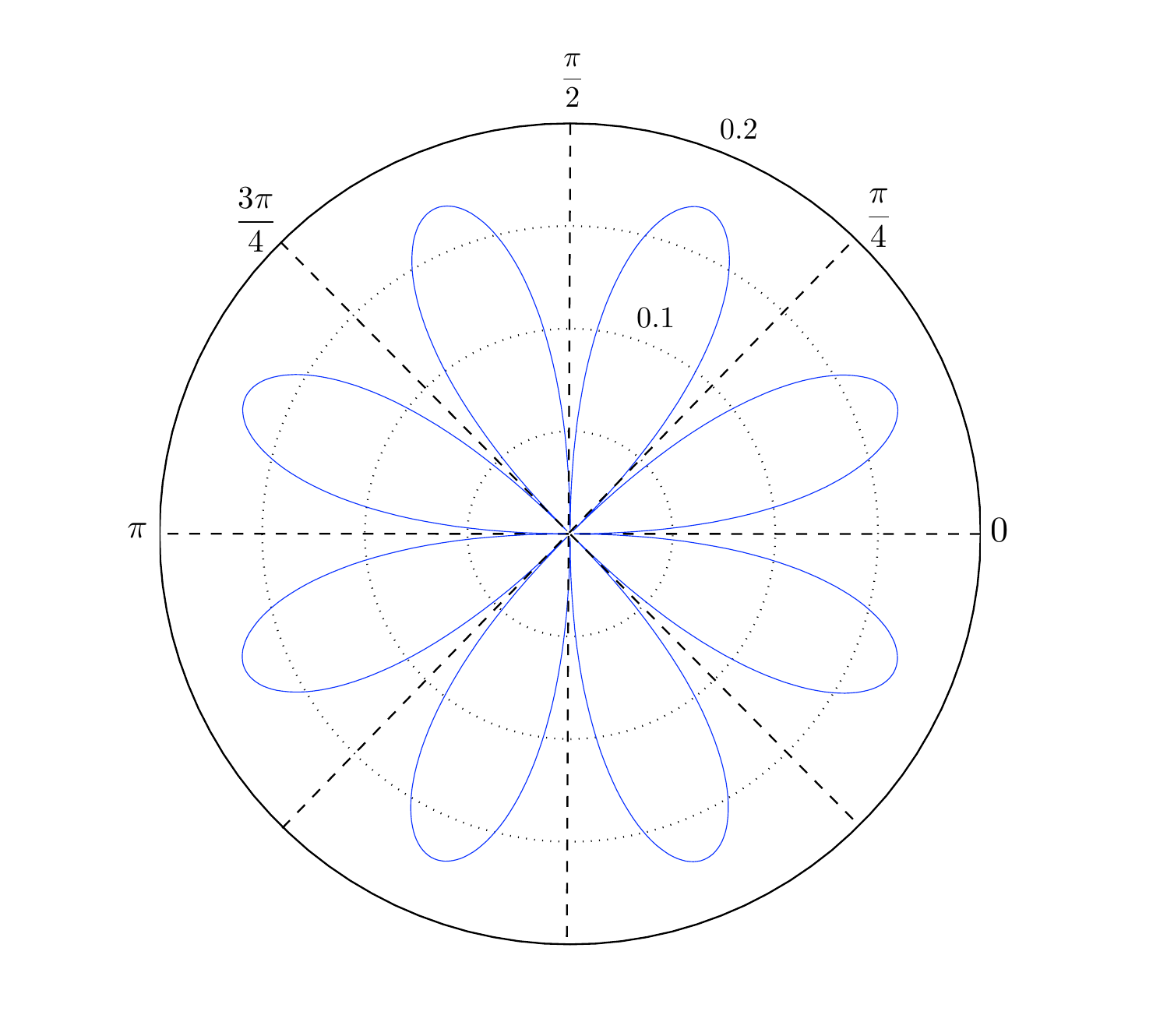}
\caption{Polar plot of the symmetric variation of $1/U(\phi)$ as a function of the filter orientation $\phi$, where $U(\phi)$ is the  bound on the elongation. The bound reaches its minimum when the orientation of the filter is exactly midway between two principal axes, whereas arbitrary elongation is achievable in the neighborhood of the four principal directions $\phi=0,\pi/4,\pi/2$ and $3\pi/4$.}
\label{bound_elongation}
\end{figure}

\begin{proposition}
\label{ellipticity}
\textit{For every $\phi$ in $[0 ,\pi)$, there exists a scale-vector $\a$ such that $\theta_{\a}=\phi$. There is however a finite bound on the elongation, and is given by
\begin{equation}
\label{bound} 
\sup \ \varrho_{\a}< U(\phi)=\frac{1+ | \nu_{\phi} |+\sqrt{1+\nu_{\phi}^2}}{1+ | \nu_{\phi}|-\sqrt{1+\nu_{\phi}^2}},
\end{equation}
where $\nu_{\phi}=\frac{1}{2}(\tan \phi-\cot \phi)\mathrm{sign}\big(\frac{\pi}{2}-\phi\big)$. The supremum is over the set of $\a$ for which $\theta_{\a}=\phi$.}
\end{proposition}
Fig. \ref{bound_elongation} illustrates the variation of $1/U(\phi)$ as a function of the orientation; the rationale behind showing the inverse plot is to avoid the blowups $U(\phi) \longrightarrow +\infty$ as $\phi \longrightarrow \theta_k$. In particular, a bound of $3+2\sqrt 2 \approx 5.8$ is attained along the orientations $\phi=(2k-1)\pi/8, 1 \leq k \leq 4$, exactly mid-way between two adjacent primal directions. This is perfectly reasonable since the control on the geometry of the box spline is minimal along these directions.

	 In order to specify the elliptical geometry of the box spline, we use either of the following equivalent descriptors as per convenience: 
	 
\begin{description}
\item (D1) Size, elongation and orientation $(s,\varrho,\theta)$.
\item  (D2) Length of the major and minor axes, and the orientation $(\sqrt \lambda_{\max},\sqrt \lambda_{\min},\theta)$.
\item (D3) Covariance matrix $\C.$  
\end{description}
		
	Descriptor (D1) stipulates the lengths of the major and minor axes as $\lambda_{\max}=s\varrho/(1+\varrho)$ and $\lambda_{\min}=s/(1+\varrho)$, respectively, whereas, (D2) gives the corresponding covariance as
\begin{equation*}
\C=\Bigg(\begin{array}{ccc} \lambda_{\max}\cos^2\theta+  \lambda_{\min}\sin^2\theta &  \frac{1}{2} \left(\lambda_{\max}-\lambda_{\min}\right)\sin2\theta  \\  \frac{1}{2}\left(\lambda_{\max}-\lambda_{\min}\right) \sin2\theta  & \lambda_{\min}\cos^2\theta+  \lambda_{\max}\sin^2\theta\end{array}\Bigg).
\end{equation*}

\subsubsection{Optimal scale-vector for a given anisotropy}
\label{optimization}

Since the covariance matrix of the four-directional box spline (cf. Eqn. \eqref{covariance}) is given by 
\begin{equation}
\label{covariance_ZP}
\C_{\a}=\frac{1}{24} \left(\begin{array}{cc} 2a^2_1+a^2_2+a^2_4 & a^2_2-a^2_4 \\
a^2_2-a^2_4 &  2a^2_3+a^2_2+a^2_4\end{array}\right),
\end{equation}
the inverse problem is that of specifying a scale-vector $\a$ such that $\C_{\a}=\C$. By introducing the positive vector $\p=(a^2_1,a^2_2,a^2_3,a^2_4),$ the problem can be reformulated as: find $\p > 0$, such that $\M\p=\c$, where  
\begin{equation*}
\M=\left(\begin{array}{cccc}2 & 1 & 0 & 1 \\ 0 & 1 & 0 & -1 \\ 0 & 1 & 2 & 1\end{array}\right), \quad \mbox{and} \quad \c=24\Big(\C(1,1), \  \C(1,2), \ \C(2,2)\Big).
\end{equation*}
The scale-vector solution is then given by $\a=\sqrt{\p}.$ As far as existence of solutions is concerned, proposition \ref{ellipticity} ensures that the linear system $\M \p=\c, \p>0$, corresponding to a given geometry $(\lambda_{\min},\lambda_{\max},\theta)$, is always solvable provided that the elongation $\varrho<U(\theta)$. Moreover, as it turns out, the linear system is under-determined and has infinitely many solutions. The idea then would be to use a scale-vector that is ``optimal'' in some sense. But first, we try to characterize the solution space of the system $\M \p=\c, \p \geq   \varepsilon \1$. For reasons that will soon be obvious, we propose to modify the positivity constraint as $\p \geq   \varepsilon \1$, where $\varepsilon$ is some arbitrarily small positive number. We observe that $\M$ is full-rank, and hence the null-space is of dimension $4-3=1$. In particular, this signifies that the solutions of $\M \p=\c$ lie on the affine subspace $\{\bar{\p}+t\e : t \in \mathbf{R} \}$, where $\bar{\p}$ is a particular solution ($\M \bar{\p}=\c$)  and $\e$ is in the null-space ($\M \e=\0$). Moreover, one can easily verify that in order to adhere to the positivity constraint, $\bar{\p}+t\e=(\bar{p}_1+t,\bar{p}_2-t,\bar{p}_3+t,\bar{p}_4-t) \geq   \varepsilon\1$, it is both necessary and sufficient that $t$ lies in the closed interval $[t_{\ell},t_r]$, where $t_{\ell}=\max(-\bar{p}_1+\varepsilon,-\bar{p}_3+\varepsilon)$ and $t_r=\min(\bar{p}_2-\varepsilon,\bar{p}_4-\varepsilon)$. In particular, we set $\e=(1,-1,1,-1)$ which is in the null-space of $\M$. Note that one can easily compute $\bar{\p}$ by pivoting one of its components and solving for the remaining three; since $\M$ is of full-rank, the reduced system is always solvable. We now use the available degree of freedom to select a solution that maximizes a certain measure of Gaussianity.

\begin{figure}
\centering
\includegraphics[width=1.0\linewidth]{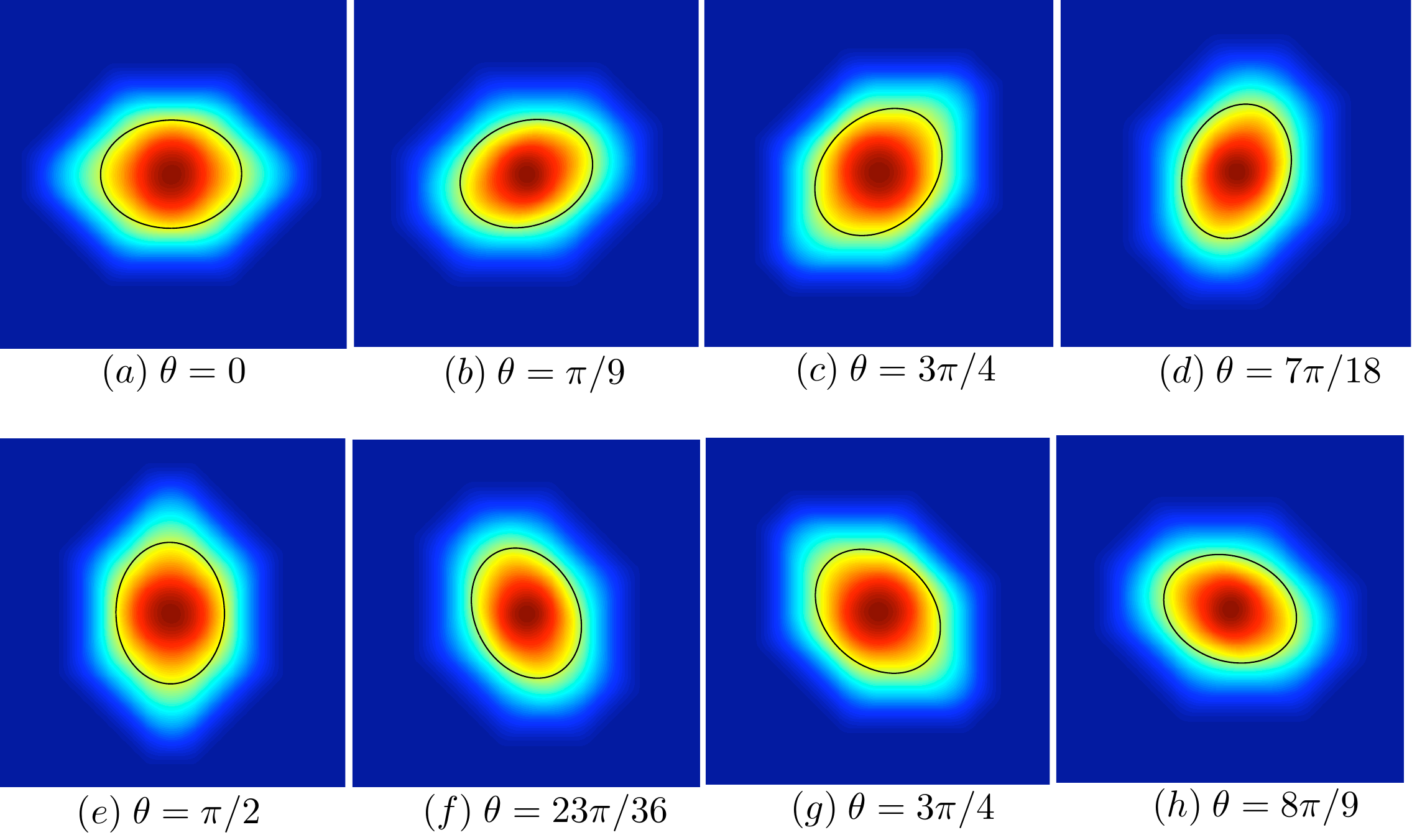}
\caption{Intensity distributions of the four-directional box splines of identical size ($s=1$) and elongation ($\varrho=2.5$), but with different orientations. The ellipse in each figure represents a level-set of the Gaussian having the same covariance as the corresponding box spline.}
\label{RUBF}
\end{figure}

	 A classical measure of the Gaussianity of a $1$-D function is its kurtosis (the fourth-order cumulant). For a centered function $f(x)$, this is defined as $\kappa=\mu_4-3\mu_2^2$, where $\mu_4$ and $\mu_2$ are the fourth-order and second-order moments of $f(x)$, respectively. The central property of this measure is that $\kappa=0$ for a true Gaussian function, and as a result, the absolute value of the kurtosis provides a measure of Gaussianity of the function. In particular, smaller absolute values correspond to more Gaussian-like functions.

	\revu{As for a bivariate function $f(\x)$, we shall use the following matrix-valued extension
\begin{equation}
\label{kurtosis_matrix}
\K=\L-\mathrm{tr}(\C)\C-2\C^2,
\end{equation}
\Rev{where $\C=\int (\x \x^T) f(\x)d\x$ and $\L=\int (\x \x^T)^2 f(\x)d\x$ are the second-order and fourth-order moment matrices} of $f(\x)$, respectively \cite{kurtosis}.} This constitutes a valid extension of the $1$-D kurtosis since \eqref{kurtosis_matrix} reduces to $\kappa=\mu_4-3\mu_2^2$ when $d=1$. Moreover, we also have the following desirable properties:\newline 

\noindent (i) If $f(\x)$ is a multivariate Gaussian, then $\K = \0$ (cf. \cite{kurtosis} for a proof). 

\noindent (ii) The Frobenius norm of $\K$, namely,
\begin{equation*}
\norm{\K} =\Big(\sum_{i,j} \abs{\K(i,j)}^2\Big)^{1/2}
\end{equation*}
is rotation-invariant, i.e., the kurtosis matrices of the rotations of $f(\x)$ have the same Frobenius norm (proof in \S\ref{proof_kurtosis}). 

 Following the above arguments, we propose to solve the optimization problem 
\begin{equation}
\label{opt1}
\p_0=\mathrm{argmin}_{\p} \ \lVert \K_{\p} \rVert^2 , \qquad  \M \p=\c , \ \p \geq   \varepsilon\1.
\end{equation}
This yields the optimal scale-vector $\a_0=\sqrt{\p_0}$ corresponding to the most Gaussian-like box spline. The rotation-invariance property ensures that the box splines of identical size and elongation but different orientation, obtained via the solutions of the above optimization problem, are as homogenous as possible. 
	
	The norm of the kurtosis matrix of  $\bbeta_{\a}(\x)$ turns out to be $\lVert \K_{\p} \rVert^2=\sum_k p^4_k + (p^2_1+p^2_3)(p^2_2+p^2_4)$ (see Appendix \S\ref{compute_kurtosis}). Substituting $p_k=\bar{p}_k+te_k$ into this expression, we arrive at the quartic polynomial 
\begin{equation*}
\zeta(t)=\sum_k (\bar{p}_k+e_k t)^4+\left\{ (\bar{p}_1+t)^2+(\bar{p}_3+t)^2 \right\} \left\{ (\bar{p}_2-t)^2+(\bar{p}_4-t)^2 \right\},
\end{equation*}
which, together with the parameterization $\p=\bar{\p}+t\e$, simplifies the problem to one of finding
\begin{equation}
\label{opt2}
t_0= \mathrm{arg \ min}_t  \ \zeta(t),  \quad t \in [t_{\ell},t_r].
\end{equation}
The optimal solution is then given by $\a_0=\sqrt{\bar{\p} +t_0\e}$. This problem however is easily solved, since the minimum is attained either at one of the interior points $(t_{\ell}, t_r)$ where $\zeta'(t)=0$, or at one of the boundary points.  In particular, we have the following simple algorithm for designing optimized Gaussian-like box splines of a specified covariance: \newline

\noindent (i) Set $p_4=1$, and compute $\bar{\p}$ by solving the sytem $\M \bar{\p}=\c.$

\noindent  (ii) Use $\bar{\p}$ to compute  $t_{\ell},t_r$ and the coefficients of $\zeta'(t)$.

\noindent  (iii) Find the real roots of $\zeta'(t)=0$ over the interval $(t_{\ell},t_r)$; denote the set of real roots by $R$. Then $\a_0=(\bar{\p}+t_0 \e)^{1/2}$, where\footnote{The tie is randomly broken if $\zeta(t)$ has multiple minimizers over $[t_{\ell},t_r]$ (this was rarely reported in practice).} $t_0=\mathrm{arg min}_t \ \zeta(t), t \in R \cup \{t_{\ell},t_r\}$. \newline

In particular, the coefficients of the cubic equation $\zeta'(t)=\zeta_1 t^3+\zeta_2t^2+\zeta_3t+\zeta_4=0$ in (iii) are given by $\zeta_1=32, \zeta_2=24(\bar p_1-\bar p_2+\bar p_3-\bar p_4), \zeta_3=16\sum \bar p_k^2 -8(\bar p_1+\bar p_3)(\bar p_2+\bar p_4), \zeta_4=4(\bar p^3_1-\bar p^3_2+\bar p^3_3-\bar p^3_4)+2(\bar p_1+\bar p_3)(\bar p^2_2+\bar p^2_4)-2(\bar p_2+\bar p_4)(\bar p^2_1+\bar p^2_3)$.

\begin{figure}
\centering
\includegraphics[width=0.75\linewidth]{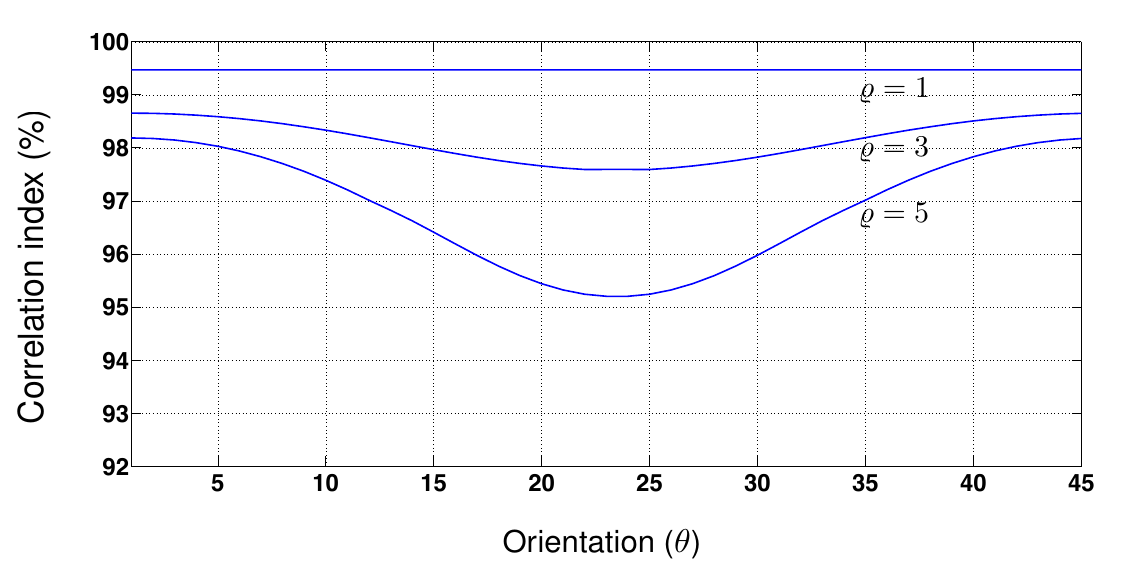}
\caption{Normalized correlation between the optimal four-directional box spline and the target Gaussian at different elongations and orientations. For a fixed elongation, the correlation is minimum at the critical orientation $\theta=22.5^{\circ}$, and improves symmetrically as $\theta$ approaches the principal orientations (cf. Fig. \ref{bound_elongation}). }
\label{errorplots}
\end{figure}

\revu{The box splines obtained using the above optimization at various orientations are shown in Fig. \ref{RUBF}. The quality of the Gaussian approximation under different practical settings of the orientation and the elongation is quantified in Fig. \ref{errorplots}.}	

	The correspondences $(1,\varrho,\theta) \leftrightarrow (a_1,a_2,a_3,a_4)$ for $0 < \theta < \pi$ and $1 \leq  \varrho <  U(\theta)$ can be pre-computed and stored in a look-up table. Note that for a given $\varrho$, the set of correspondences $(1,\varrho,\theta) \leftrightarrow (a_1,a_2,a_3,a_4)$ have an inherent four-fold symmetry in $\theta$ owing to the presence of the four principal directions. Hence, it suffices to store the scale-vector correspondences for $0 < \theta < \pi/4$ which reduces the size of the LUT by a factor of four. Indeed, for any arbitrary size $s>1$, orientations $0 < \theta <\pi,$ and elongation $1 \leq  \varrho <  U(\theta)$, the corresponding scale-vector is then obtained through the following operations:

\noindent (O1) Rotation:
\begin{equation*}
\phi=    \begin{cases}
    \theta & \mbox{for} \quad 0 < \theta<\pi/4\\
    \theta-\pi/4 & \mbox{for} \quad \pi/4 < \theta<\pi/2\\
     \theta-\pi/2 & \mbox{for} \quad \pi/2 < \theta<3\pi/4\\
     \theta-3\pi/4& \mbox{for} \quad 3\pi/4 < \theta<\pi. \\
    \end{cases}
 \end{equation*}
 
 \noindent (O2) Find $(a_1,a_2,a_3,a_4)$ corresponding to $(1,\varrho,\phi)$ using the LUT. The desired scale-vector is then given by the following permutation and rescaling:
 \begin{equation*}
 (a_1,a_2,a_3,a_4)\mapsto     \begin{cases}
    \sqrt s(a_1,a_2,a_3,a_4) &  \mbox{for} \quad 0< \theta < \pi/4\\
    \sqrt s (a_2,a_3,a_4,a_1) &  \mbox{for} \quad \pi/4< \theta< \pi/2 \\
    \sqrt s (a_3,a_4,a_1,a_2) & \mbox{for} \quad \pi/2< \theta< 3\pi/4\\
    \sqrt s (a_4,a_1,a_2,a_3)& \mbox{for} \quad  3\pi/4< \theta<\pi. \\
 \end{cases}
\end{equation*}
\vspace{0.1in}

\subsection{Higher-order box splines}
\label{discussion}

\begin{figure}
\centering
\includegraphics[width=1.0\linewidth]{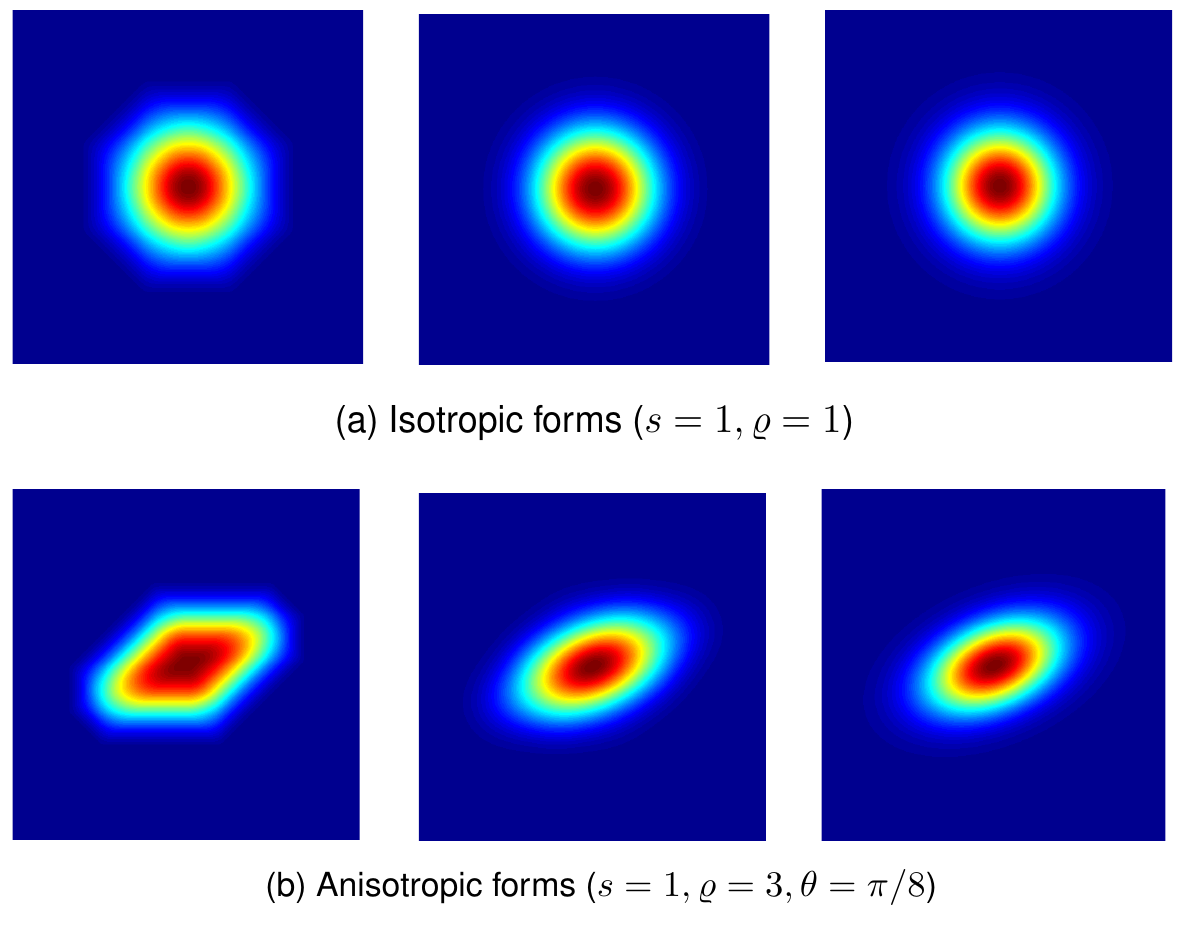}
\caption{Higher-order box splines through iterative convolutions. \textbf{Left}: The reference four-directional box spline; \textbf{Center}: Iterated box spline obtained by convolving the (rescaled) four-directional box spline with itself; \textbf{Right}: Target Gaussian having identical covariance.}
\label{iterated_box_splines}
\end{figure}

	As suggested by the convergence result \eqref{conv1}, the Gaussian-like nature of  the four-directional box splines can be improved by using more directions. Implementing the corresponding space-variant filtering using the algorithm in \S\ref{2Dalgo} however turns out to be challenging and not very practical---the principal axes of these box splines are generally along off-grid directions, and one needs to interpolate the image for implementing the associated running-sums. 
	
	The iterated four-directional box splines $\beta^{4,m}_{\a}(\x)$ introduced in \S\ref{convergence} provide a practical alternative. These box splines rapidly converge to a Gaussian with the increase in $m$. Also, note that the four-directional box spline and its iterates have identical covariances. This implies that the algorithm in \S\ref{optimization} can be used for optimizing the iterated box splines as well. The first two iterates of the four-directional box spline along with the target Gaussian are shown in Fig. \ref{iterated_box_splines}. It is seen that $\beta^{4,2}_{\a}(\x)$ resembles the target Gaussian very closely. In fact, the minimum correlation coefficient rises from $95\%$ to $99\%$ for $m=2$ (cf. Fig. \ref{errorplots}). In practice, we can thus implement a higher-order Gaussian-like filtering by simple iterations of the algorithm in \S \ref{filtering}. It suffices to set the scale-vector in the algorithm as $\a/\sqrt{m}$, where $m$ is the number of iterations.

\section{EXPERIMENTAL RESULTS}
\label{VII}

\subsection{Computation time}

	\revu{The space-variant filtering using the four-directional box spline was implemented in Java on a $2.66$ GHz Intel system. The typical execution times required for convolving a $512 \times 512$ image with kernels of various sizes are shown in Table \ref{computation_time}. It is clear that the run time is independent of the size of the kernel.}

\begin{table}[!htbp]
\caption{Average computation time for box splines of different sizes.} 
\label{computation_time} 
\centering
\begin{tabular}{|c|ccccc|}
\hline
Size ($s$)  & 1  &  2   &  4   &  8  & 16   \\
\hline
Time (millisec.) & 101   &   100    &   103   &   101  & 100 \\
\hline
\end{tabular}
\end{table}

\subsection{Application: Feature-preserving smoothing}

	\revu{We now present an application \Rev{to demonstrate the space-variant algorithm} described in \S\ref{filtering}. Filtering of noisy images using isotropic Gaussian filters often results in excessive blurring of the anisotropic image features. Diffusion filters are known to perform better in such cases \cite{Weikert1996}. As an alternative, we propose to filter the corrupted image using our anisotropic Gaussian-like filters, where we adapt the size, elongation and orientation of the filter to the local image features.} The main idea is to locally average the image using elliptical windows that have been elongated along the image feature (orthogonal to the local gradient). This induces more smoothing along the direction of minimal intensity variation resulting in the suppression of the ambient noise, while preserving the sharpness of the image features.

		To derive an estimate of the local image anisotropy, we use the paradigm of \textit{structure tensors} \cite{structure_tensor}, where the local orientation $\theta(\x)$ is estimated through the minimization of a certain weighted norm of the directional derivative. In particular, if we denote the directional derivative of $f(\x)$ along along $\u_{\theta}=(\cos\theta,\sin\theta)$ by $D_{\theta}f(\x)$, then $\theta(\x)$ is given by the minimizer of 
\begin{equation}
\label{ST_prob}
\int_{\Omega} w(\s) \abs{(D_{\theta}f)(\x-\s)}^2  d\s,
\end{equation}
where $\Omega$ is the support of the isotropic averaging window $w(\x)$. The solubility of the above optimization problem follows from the observation that this can be recast as an eigenvalue problem. In particular, by expressing the directional derivative in terms of the gradient $\g(\x)$, namely as $D_{\theta}f(\x)=\u_{\theta}^T \g(\x)$, one can rewrite \eqref{ST_prob} as
\begin{equation}
\label{eigen}
  \u_{\theta}^T \J(\x) \u_{\theta},
\end{equation}
where the structure tensor $\J(\x)$ is the $2 \times 2$ positive-definite matrix 
\begin{equation*}
\int_{\Omega} w(\s) \big(\g \g^T\big)(\x-\s) d\s.
\end{equation*}
The local orientation $\theta^{\star}(\x)$ is then given by the minimizer of \eqref{eigen} associated with the minimum eigenvalue of $\J(\x)$.

\begin{figure}
\centering
\includegraphics[width=1.0\linewidth]{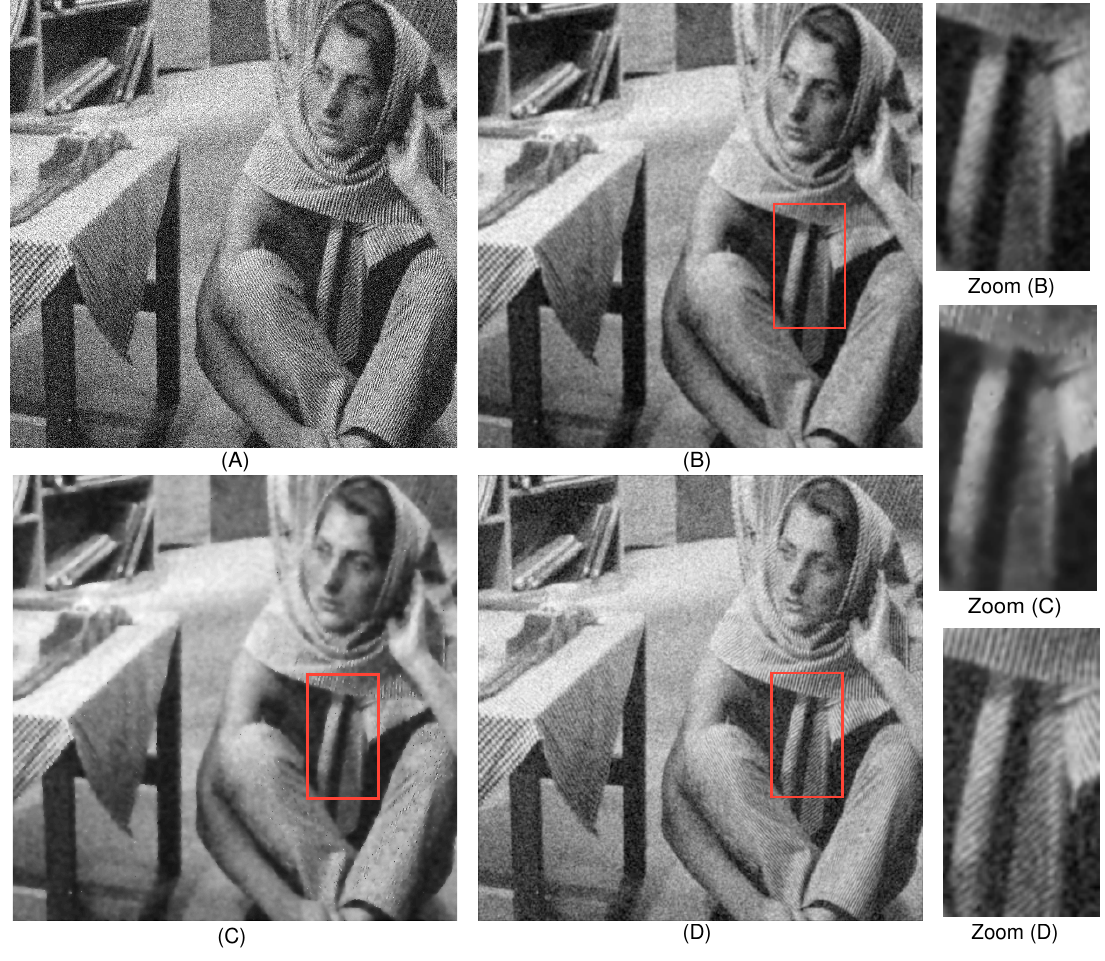} 
\caption{Results on a test image. (\textbf{A}) \textit{Barbara} corrupted with additive Gaussian noise, $\mathrm{PSNR}=18.0 \text{ dB}$; (\textbf{B}) Isotropic smoothing, $\mathrm{PSNR}=23.10 \text{ dB}$; (\textbf{C})  Diffusion filtering, $\mathrm{PSNR}=23.25 \text{ dB}$; (\textbf{D}) Our algorithm, $\mathrm{PSNR}=23.58 \text{ dB}$.}
\label{Barbara}
\end{figure}

\begin{figure}
\centering
\includegraphics[width=1.0\linewidth]{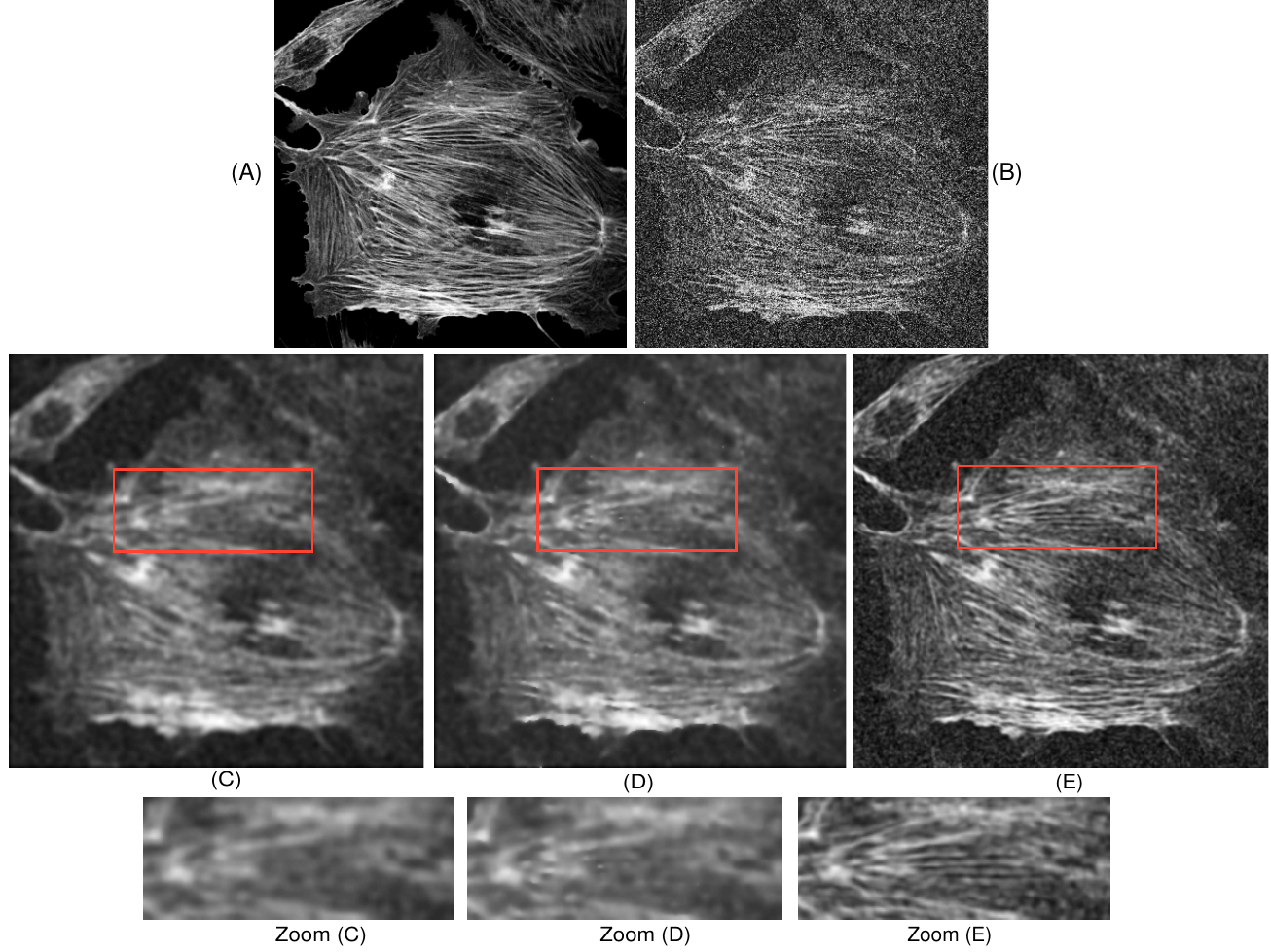} 
\caption{Results on a real image. (\textbf{A}) Noise-free immunofluorescence image of actin fibres (Courtesy of C. Aemisegger, CMIA, University of Z�rich); (\textbf{B}) Image corrupted with additive Gaussian noise, $\mathrm{PSNR}=12.20 \ \mathrm{dB}$; (\textbf{C}) Isotropic smoothing, $\mathrm{PSNR}=15.38 \ \mathrm{dB}$; (\textbf{D}) Diffusion filtering, $\mathrm{PSNR}=15.50 \ \mathrm{dB}$; (\textbf{E}) Our algorithm, $\mathrm{PSNR}=15.80 \ \mathrm{dB}$.}
\label{Fibres}
\end{figure}

	In view of the definitions in  \eqref{def_eigen} and the fact that the eigenvalues of $\J(\x)$ are always non-negative, we propose to estimate the elongation as follows: We set $\varrho^{\star}=\lambda_{\max}/\lambda_{\min}$ if both eigenvalues are non-zero, equal to $1$ if both are zero (locally isotropic intensity), and equal to $\max(1,\lambda)$ if only one of the eigenvalues $\lambda$ is non-zero. Finally, we estimate the size of the box spline as $s^{\star}=\lambda_{\max}+\lambda_{\min}$. The triple $(s^{\star},\varrho^{\star},\theta^{\star})$ is then used to compute the optimal scale-vector using the algorithm described in \S\ref{geometry_control}. The components of $\J$ can be efficiently computed using simple convolution and pointwise operations; we refer the reader to \cite[Chapter 13]{structure_tensor} for implementation details. The  main steps of the proposed smoothing algorithm are:
	
\begin{itemize}
\item Computation of the structure-tensor.

\item Pre-integration of  the corrupted image using the running-sum filters.
\item Computation of the triple $(s^{\star},\varrho^{\star},\theta^{\star})$ at every feature location using the structure tensor. This is used to compute the scale-vector $\a(\n)$ of the optimal Gaussian-like box spline using the algorithm in \S\ref{geometry_control}. Isotropic box splines are used in the uniform-intensity regions; we set $\a(\n)=(\sigma ,\sigma ,\sigma ,\sigma )$, where $\sigma$ is proportional to the noise variance.
\item Computation of the FD mesh using $\a(\n)$, and its application to the pre-integrated image.
\end{itemize}

	 To demonstrate the effectiveness of our strategy in preserving oriented patterns in noisy images, we compare the results obtained from our algorithm with those obtained using  the (fixed-scale) isotropic Gaussian filter and the Perona-Malik diffusion filter \cite{PeronaMalik}. We use the standard test image of \textit{Barbara} and corrupt it with additive Gaussian noise. The variance of the noise is used to set the size of the Gaussian for the isotropic smoothing. \Rev{The parameters used for the Perona-Malik filter were typical: time step of $0.1$, conductance in the range of $10 \sim 30$, and a total of $15 \sim 30$ iterations. The parameters were manually tuned to optimize the PSNR, and also to avoid blocking artifacts.} Fig. \ref{Barbara} shows the results obtained from the different filters. As far as the quantitative evaluation of the filters is concerned, our algorithm clearly outperforms both isotropic and diffusion filters in terms of the Peak-Signal-to-Noise-Ratio (PSNR). Moreover, as shown in the zoomed-in sections of the respective images, the oriented stripes on the clothes are quite faithfully restored by our algorithm. A significant amount of blurring of the stripes is seen in the results obtained using isotropic and diffusion filtering. The non-linear diffusion filter, however, tends to perform better at low PSNRs in the range of $5$-$10$ dB (cf. Table \ref{noise_levels}).
	  	  
	  Next, we compare the results on a real biological image and at a much lower PSNR of  around $12$ dB. We consider the fluorescence image shown in Fig. \ref{Fibres}, which exhibits numerous elongated fiber-like structures. The parameters of the isotropic filter and the diffusion filter are set as in the previous case, except that the iteration count for the latter is increased to $15$. As before, the improvement of the PSNR obtained using our filter is higher. Importantly, as seen from the zooms, our algorithm results in significantly less merging of the close fibers and blurring of the finer ones. The average execution time of our algorithm is $0.6$ seconds for a $512 \times 512$ image, which includes the computation of the structure-tensor, the running-sums, the optimal scale-vector, the interpolated samples and the finite-differences.

	The four-directional box splines can also be used to derive fast space-variant detectors based on Gaussian forms, e.g., the Laplacian-of-the-Gaussian (LoG) or the so-called Mexican-hat detector. We refer the interested reader to \cite{ISBI_kunal}, where the isotropic forms of the four-directional box spline were used to realize a fast and scalable Mexican-hat-like detector. In particular, a modified version of the space-variant algorithm described in \S\ref{filtering}  is used to design an efficient  coarse-to-fine strategy for the detection of centers and radii of cells/nuclei in fluorescence images.

\begin{table}[!t]
\caption{Comparison of the filters at different noise levels using the test image of \textit{Barbara}. The table shows the PSNR of the outputs.} 
\label{noise_levels} \vspace{1mm}
\centering
\begin{tabular}{|c|cccccc|}
\hline
Input PSNR (dB) &     10.0  &  	    12.0   &      14.0   &       16.0  &        18.0 &         20.0   \\
\hline
Isotropic filter &   15.38   &    18.20    &   20.20   &    21.65  &     23.10 &     24.30  \\
\hline 
Diffusion filter &  \bf{15.48}   &    18.31    &    20.30   &    21.70   &   23.25 &     24.35 \\
\hline
Our filter & 	  15.45   &     \bf{18.38}    &   \bf{20.57}   &   \bf{21.94}  &    \bf{23.58} &     \bf{24.56} \\
\hline
\end{tabular}
\end{table}

\section{CONCLUSION}
\label{VIII}

	In this paper, we presented a framework for elliptical filtering using the radially-uniform box splines. The associated space-variant filtering was efficiently realized using running-sums and local finite-differences. The attractive features of our algorithm are:
\begin{itemize}
\item  the $O(1)$ computational complexity per pixel, and
\item the use of real-valued parameters for continuously controlling the shape and size of the filter.
\end{itemize}
Our filtering paradigm offers a nice trade-off between the quality of approximation of Gaussians and the computational complexity of linear space-variant filtering.
	
	We also presented a closed form solution for the problem of constructing four-directional box splines with given covariances. The scope of our algorithm was demonstrated through the realization of a smoothing filter that can adapt to the local image characteristics.

\section{APPENDIX}
\label{derivations}

\subsection{Proof of theorem \ref{convergenceOrder}}
\label{appendix A}

	We first establish that the Fourier sequence $\h{\bbeta}^2_{\a(2)}(\bw),\h{\bbeta}^3_{\a(3)}(\bw),\ldots$ convergences  pointwise to a Gaussian:
\begin{equation}
\label{conv}
\lim_{N \longrightarrow \infty} \h{\bbeta}^{N}_{\a(N)}(\bw)= \exp\Big(-\frac{\sigma^2}{2}\norm{ \bw}^2\Big).
\end{equation}	
We then show that the above convergence is also in the $\mathbf{L}^2(\R^2)$ norm. This will establish the theorem, since it is well-known that the Fourier transform of a Gaussian is a Gaussian, and that $f_n \longrightarrow g$ in $\mathbf{L}^2$ if $\hat{f}_n \longrightarrow \hat{g}$ in $\mathbf{L}^2$.

	To derive \eqref{conv}, we note that $\h{\varphi}_{a,\theta}(\bw)= \h{\beta}_a(\u_{\theta}^T\bw) = \sinc \left(a \u_{\theta}^T\bw/2\right)$, where $\sinc(x)=\sin(x)/x$ for $x \neq 0$, and equals $1$ at the origin. Then, the convolution-multiplication rule gives
 \begin{equation}
\label{A1}
\h{\bbeta}^{N}_{\a(N)}(\bw)= \prod_{k=1}^N \h{\varphi}_{a_k(N),\theta_k}(\bw) =\prod_{k=1}^N \sinc\left(\frac{a_k(N)}{2}\u_{\theta_k}^T\bw\right). 
\end{equation}
Using the estimate $\sinc(x)=1-x^2/6+O(x^4)$ for $\abs{x}<1$, and substituting $a_k(N)=\sigma \surd(24/N)$ into \eqref{A1}, we have 
\begin{equation}
\label{exp}
\h{\bbeta}^{N}_{\a(N)}(\bw)= \prod_{k=1}^N \left(1-\frac{\sigma^2}{N}(\u_{\theta_k}^T\bw)^2+O\left(N^{-2}\right)\right) \qquad (\norm{\bw} < cN)
\end{equation}
where $c$ is some positive constant. By developing the quadratic factor $(\u_{\theta_k}^T\bw)^2$ and the product in \eqref{exp}, we arrive at the estimate
\begin{align}
\label{pre_conv}
\h{\bbeta}^{N}_{\a(N)}(\bw)&=  \prod_{k=1}^N \left\{1- \frac{\sigma^2}{2N}\norm{\bw}^2+\frac{\sigma^2}{2N}(\w_1^2-\w_2^2) \cos 2\theta_k+\frac{\sigma^2}{N}\w_1\w_2 \sin2\theta_k+O\left(N^{-2}\right)\right\} \nonumber \\
&=   \Big(1-\frac{\sigma^2 }{2N}\norm{\bw}^2\Big)^N +\frac{\sigma^2}{2N} (\w_1^2-\w_2^2)\left(1-\frac{\sigma^2 }{2N}\norm{\bw}^2\right)^{N-1} \sum_{k=1}^N \cos 2\theta_k  \nonumber \\
&+ \frac{\sigma^2}{N} \w_1\w_2\left(1-\frac{\sigma^2 }{2N}\norm{\bw}^2\right)^{N-1} \sum_{k=1}^N \sin 2\theta_k+O\left(N^{-2}\right) \nonumber \\
&=   \Big(1-\frac{\sigma^2 }{2N}\norm{\bw}^2\Big)^N +O\left(N^{-2}\right) \qquad (\norm{\bw} < cN).   
\end{align}
This is exactly where the fact that $\theta_k$ are uniformly distributed over $[0,\pi)$ is invoked: the cancellation of the linear factors in the second step is based on the identities $\sum_{k=1}^N \cos2\theta_k=0$, and $\sum_{k=1}^N \sin 2\theta_k=0$, where $\theta_k=(k-1)\pi/N$. Since $(1-x/m)^m$ converges to $\exp(-x)$ as $m \longrightarrow \infty$, it can now be readily seen that \eqref{conv} follows as the limiting case of \eqref{pre_conv} .

To demonstrate that \eqref{conv} holds in the $\mathbf{L}^2$ norm sense, it suffices to show the sequence of error functions $\mathcal{E}_N(\bw)=\h{\bbeta}^{N}_{\a(N)}(\bw)-\exp(-\sigma^2 \norm{ \bw}^2/2)$ converge to zero in the above norm, i.e., $\norm{\mathcal{E}_N}_{\mathbf{L}^2} \longrightarrow 0$ as $N \longrightarrow \infty$. Since we have already demonstrated that $\mathcal{E}_N(\bw) \longrightarrow 0$ pointwise, all we need to show in order to invoke the dominated convergence theorem is that the sequence $\abs{\mathcal{E}_2(\bw)}, \abs{\mathcal{E}_3(\bw)}, \ldots$ is uniformly bounded by a $\mathbf{L}^2$ function. Moreover, since
\begin{equation*}
| \mathcal{E}_N(\bw)| \leq | \h{\bbeta}^{N}_{\a(N)}(\bw)|+\exp(-\sigma^2 \norm{\bw}^2/2),
\end{equation*}
it, in fact, suffices to show that each $| \h{\bbeta}^{N}_{\a(N)}(\bw)|$ admits such a bound. 

		The main idea behind establishing such a bound is that the above mentioned sequence can be covered by a Gaussian in a neighborhood of the origin and by a function with sufficient decay at the tails, both of which are independent of $N$. Indeed, using the estimate $\sinc(u)\leq  1-u^2/\pi^2$ for $\abs{u} \leq  \pi$, one can verify that 
\begin{align*}
\label{B1}
\abs{\h{\bbeta}^{N}_{\a(N)}(\bw)}& = \prod_{k=1}^N\Big| \sinc\Big(\frac{\sqrt 6\sigma}{\sqrt N}\u_{\theta_k}^T\bw\Big)\Big|   \nonumber \\
&\leq    \prod_{k=1}^N\Big(1-\frac{6\sigma^2 \abs{\u_{\theta_k}^T\bw}^2 }{\pi^2 N}\Big) \nonumber \\
&\leq   \exp\big(-C_1 \norm{\bw}^2\big) \qquad (\norm{\bw}  <   \delta)
\end{align*}
As far as the tail is concerned, the Cauchy-Schwarz inequality, $| \u_{\theta_k}^T\bw| \leq  \lVert \u_{\theta_k} \rVert \cdot \lVert \bw \rVert= \lVert \bw \rVert$, gives
\begin{equation*}
\label{B2}
\big|\h{\bbeta}^{N}_{\a(N)}(\bw)\big| = \prod_{k=1}^N\Big| \sinc\Big(\frac{\sqrt 6\sigma}{\sqrt N}\u_{\theta_k}^T\bw\Big)\Big| \leq  \frac{C_2}{\norm{\bw}^2} \ \ \ (\norm{\bw}  \geq   \delta).
\end{equation*}
Here $C_1,C_2$ and $\delta$ are appropriate positive constants that are independent of $N$. Combining the above estimates, we see that
\begin{equation*}
\label{right}
\big |\h{\bbeta}_{\a(N)}^N(\bw)\big | \leq \exp\big(-C_1\norm{\bw}^2\big)+\frac{C_2}{\norm{\bw}^{2}} \left[1-\rect\left(\frac{\norm{\bw}}{\delta}\right)\right]
\end{equation*}
for all $\bw$. Since the function on the right is indeed in $\mathbf{L}^2(\mathbf{R}^2)$, this establishes the desired bound, and consequently, the norm convergence.

\subsection{Covariance matrix}
\label{appendix B}

	We begin with the observation that if $f(\x)$ and $g(\x)$ are symmetric (about the origin) and have a total mass of unity, then $\C_{f \ast g}=\C_f+\C_g$, where $\C_f$ denotes the covariance matrix of $f(\x)$. Indeed, by noting that $\hat{f}(\0)=\hat{g}(\0)=1$ (unit mass) and $\partial_i \hat{f}(\0)=\partial_i \hat{g} (\0)=0$ (by symmetry), and by recalling the multiplication-differentiation rule $\int x_i x_j f(\x)  d\x = -\partial_i \partial_j \hat{f}(\0)$, we have
\begin{align*}
\C_{f \ast g}(i,j)&=\int x_i x_j (f \ast g)(\x) d \x \nonumber \\
&=   - \hat{g}(\0) \partial_i \partial_j \hat{f}(\0) -\hat{f}(\0) \partial_i \partial_j \hat{g} (\0)  - \partial_i \hat{f}(\0) \partial_j \hat{g}(\0)   - \partial_i \hat{g}(\0) \partial_j \hat{f}(\0)  \nonumber \\
&= - \partial_i \partial_j \hat{f}(\0)  - \partial_i \partial_j \hat{g}(\0) \\
&= \C_f(i,j)+\C_g(i,j).
\end{align*}
 Since the directional box distributions $\varphi_{a_k,\theta_k}(\x)$ satisfy these criteria, we have that  $\C^N_{\a}=\sum_k \C_k$, where $\C_k$ is the covariance matrix of $\varphi_{a_k,\theta_k}(\x)$. We explicitly compute the component $\C(1,2)$; the remaining components can be similarly derived. Using the multiplication-differentiation rule again, we have
\begin{align*}
\C_k(1,2)&=\int x_1 x_2 \varphi_{a_k,\theta_k}(\x)   d \x  \\
&=  -\partial_1 \partial_2 \hat{\beta}_{a_k}(\u_{\theta_k}^T\bw)\Big |_{\bw=\0} \\
&=\frac{a_k^2}{24}  \sin 2\theta_k.
\end{align*}
Therefore, $\C^N_{\a}(1,2)=\sum_k \C_k(1,2)=\sum_k a_k^2 \sin 2\theta_k /24$. 
	
	As far as the positive-definite nature of $\C^N_{\a}$ is concerned, it will suffice to show that its eigenvalues
\begin{equation*}
\lambda_{\max}=\frac{1}{2}\left(\sum a_k^2 + \sqrt D\right) \ \text{ and } \ \lambda_{\min}=\frac{1}{2}\left(\sum a_k^2 - \sqrt D\right)
\end{equation*}
where $D=(\sum a^2_k \cos(2\theta_k))^2+(\sum a^2_k \sin(2\theta_k))^2$, are strictly positive. This is obviously the case for $\lambda_{\max}$. Moreover, the inequality
\begin{align*}
\Big(\sum_k a_k^2\Big)^2 - D&=\Big(\sum_k a_k^2\Big)^2-\Big(\sum_{k} a^2_k \cos(2\theta_k)\Big)^2-\Big(\sum_{k} a^2_k \sin(2\theta_k)\Big)^2  \\
&= 2 \sum_{k \neq \ell} a_k^2 a_{\ell}^2 \Big(1-\cos(2\theta_k-2\theta_{\ell})\Big) > 0
\end{align*}
tells us that $\sum a_k^2 > \sqrt D$. Hence, $\lambda_{\min}$ is strictly positive as well.

\subsection{Proof of proposition \ref{ellipticity}}
\label{proof:ellipticity}

	Following definition \eqref{def_eigen}, the dependence of orientation of the box spline $\bbeta_{\a}^4(\x)$ on the scale-vector can be expressed as
\begin{equation}
\label{direction}
\tan \theta_{\a} = \nu+\mathrm{sign}(a_2-a_4) \sqrt{1+\nu^2} \qquad (0 < \theta <\pi)
\end{equation}
where $\nu=(a^2_3-a^2_1)/(a^2_2-a^2_4)$. 

	We note the following: it is both necessary and sufficient that $a_2>a_4$ (resp. $a_2<a_4$) for the box spline to be oriented between $0<\theta_{\a}<\pi/2$ (resp. $\pi/2<\theta_{\a}<\pi$), and it is the map $(a_1,a_2,a_3,a_4) \mapsto \left(\nu, \mathrm{sign}(a_2-a_4)\right)$ that uniquely determines the orientation of the box spline. Indeed, the uniqueness aspect is based on the argument that, for $0 < \theta <\pi/2$, \eqref{direction} reduces to $\tan \theta_{\a}=\nu+\sqrt{1+\nu^2}$ as a consequence of the necessary condition $a_2>a_4$. This implicitly represents a one-to-one between $\theta_{\a}$ and $\nu$ over the domains $(0,\pi/2)$ and $(-\infty,\infty)$, since the map $\theta_{\a} \mapsto \tan \theta_{\a}$ from $(0,\pi/2)$ into $(0,\infty)$, and the map $\nu \mapsto \nu+\sqrt{1+\nu^2}$ from $(-\infty,\infty)$ into $(0,\infty)$ are both strictly monotonic. In a similar vein, a one-to-one between $\theta_{\a}$ and $\nu$ over the domains $(\pi/2,\pi)$ and $(-\infty,\infty)$ can be established. In particular, we have
\begin{equation}
\label{bijections}
\nu= \frac{1}{2}(\tan \theta_{\a}-\cot \theta_{\a})\mathrm{sign}\left(\frac{\pi}{2}-\theta_{\a}\right).
\end{equation}
That is, given any orientation $\theta_{\a}=\phi$, the corresponding $\nu_{\phi}$ is uniquely specified by \eqref{bijections}. This establishes the first part of the proposition, since there trivially exists some positive vector $(a_1,\ldots,a_4)$ such that $(a^2_3-a^2_1)/(a^2_2-a^2_4)=\nu_{\phi}$.

	As far as the bound is concerned, we observe that the elongation can be expressed as  $\varrho_{\a}=1+ 2/(\gamma-1),$ where $\gamma=\sum a^2_k/\sqrt{D}_{\a}  \geq   1$. For a given orientation $\theta_{\a}=\phi$, the components of the feasible scale-vectors bear the relation $(a^2_1-a^2_3)=\nu_{\phi}(a^2_4-a^2_2)$, and thus we have that
\begin{align*}
\gamma &=\frac{\sum a^2_k}{\sqrt{(a^2_3-a^2_1)^2+(a^2_2-a^2_4)^2}}   \\
&= \frac{a^2_1+a^2_3}{\sqrt{(a^2_3-a^2_1)^2+(a^2_2-a^2_4)^2}}+\frac{a^2_2+a^2_4}{\sqrt{(a^2_3-a^2_1)^2+(a^2_2-a^2_4)^2}}  \\
&= \frac{1}{\sqrt{1+\nu_{\phi}^2}}\frac{a^2_1+a^2_3}{|a^2_1-a^2_3|}+\frac{|\nu_{\phi}|}{\sqrt{1+\nu_{\phi}^2}}\frac{a^2_2+a^2_4}{|a^2_2-a^2_4|}  > \frac{1+|\nu_{\phi}|}{\sqrt{1+\nu_{\phi}^2}} 
\end{align*}
following the trivial inequalities $a^2_1+a^2_3 > |a^2_1-a^2_3|$, and $a^2_2+a^2_4 > |a^2_2-a^2_4|$. Consequently, 
\begin{equation}
\label{SUP}
\varrho_{\a}=1+ \frac{2}{\gamma-1}< \frac{1+|\nu_{\phi}|+\sqrt{1+\nu_{\phi}^2}}{1+|\nu_{\phi}|-\sqrt{1+\nu_{\phi}^2}}.
\end{equation}
The above bound is tight since it can be approached arbitrary closely by making the scales $a_{\ell}$  and $a_k \ (\theta_{\ell} < \phi <\theta_k)$ arbitrarily large.

\subsection{Rotation-invariance}
\label{proof_kurtosis}

Let $\K$ and $\K_{\theta}$ denote the kurtosis matrices of $f(\x)$ and its rotation $f(\R^T_{\theta}\x)$, respectively, where $\R_{\theta}$ is the rotation matrix. Observe that the  matrices $\L_{\theta}$ and $\L$ are related as
\begin{align*}
\L_{\theta}=\int  (\x \x^T)^2 f(\R^T_{\theta}\x)  d \x &=\int  \R_{\theta} (\y \y^T)^2 \R^T_{\theta} f(\y)  d\y \qquad  (\y = \R^T_{\theta}\x) \nonumber \\
&=\R_{\theta} \left(\int (\y \y^T)^2 f(\y)  d \y \right) \R^T_{\theta}  \nonumber \\
&= \R_{\theta} \L \R^T_{\theta} .
\end{align*}
Similarly, we have $\C_{\theta}=\R_{\theta} \C \R^T_{\theta}$. This also gives us the equivalence $\mathrm{tr}(\C_{\theta})=\mathrm{tr}(\R_{\theta} \C \R^T_{\theta})=\mathrm{tr}(\C \R^T_{\theta} \R_{\theta})=\mathrm{tr}(\C)$ following the identities $\mathrm{tr}(\A\B)=\mathrm{tr}(\B\A)$ and $ \R^T_{\theta} \R_{\theta}=\I$. We can then write 
\begin{align*}
\K_{\theta}&= \L_{\theta}-\mathrm{tr}(\C_{\theta})\C_{\theta}-2\C_{\theta}^2  \R_{\theta} \L \R^T_{\theta} -\mathrm{tr}(\C) \R_{\theta} \C \R^T_{\theta}-2\R_{\theta} \C^2 \R^T_{\theta}  \\
&= \R_{\theta} (\L -\mathrm{tr}(\C)\C-2\C^2)\R^T_{\theta}= \R_{\theta} \K \R^T_{\theta}.
\end{align*}
Our claim follows immediately, since $\norm{\K_{\theta}}^2=\mathrm{tr}(\K_{\theta}^T \K_{\theta})=\mathrm{tr}(\R_{\theta} \K^T \K \R^T_{\theta})=\mathrm{tr}(\K^T \K)=\norm{\K}^2$.

\subsection{Kurtosis measure} 
\label{compute_kurtosis}

    In order to compute the kurtosis matrix, we only need to evaluate the fourth-order moments; the second-order moments are already known. In particular, using Fourier identities similar to the ones used in $\S$\ref{appendix B}, one can derive the following expressions:
 \begin{align*}
\int x_1^4 \bbeta_{\a}(\x)  d\x&=\frac{1}{4}\mu_4\left(4a_1^4+a_2^4+a_4^4\right)+\frac{1}{2}\mu_2^2\left(6a_1^2a_2^2+6a_1^2a_4^2+3a_2^2a_4^2\right), \nonumber \\
\int x_1^3x_2 \bbeta_{\a}(\x)  d\x&=\frac{1}{4}\mu_4\left(a_2^4-a_4^4\right)+\frac{3}{2}\mu_2^2 a_1^2\left(a_2^2-a_4^2\right), \nonumber \\
\int x_1^2x_2^2 \bbeta_{\a}(\x)  d\x&=\frac{1}{4}\mu_4\left(a_2^4+a_4^4\right)+\frac{1}{2} \mu_2^2\left( a_1^2a_2^2+a_1^2a_4^2+a_2^2a_3^2+a_3^3a_4^2-a_2^2a_4^2+2a_1^2a_3^2\right), \nonumber \\
\int x_1x^3_2 \bbeta_{\a}(\x)  d\x&=\frac{1}{4}\mu_4\left(a_2^4-a_4^4\right)+\frac{3}{2}\mu_2^2a_3^2\left(a_2^2-a_4^2\right), \nonumber \\
\int x_2^4 \bbeta_{\a}(\x)  d\x&=\frac{1}{4}\mu_4\left(4a_3^4+a_2^4+a_4^4\right)+\frac{1}{2}\mu_2^2\left(6a_2^2a_3^2+6a_3^2a_4^2+3a_2^2a_4^2\right),
\end{align*}
where $\mu_4=1/80$ and $\mu_2=1/12$ denote the fourth and second-order moments of $\beta_1(x)$, respectively. These provide the components of the matrix $\L_{\a}$, which in turn leads to the following simple expression for the kurtosis matrix 
\begin{equation}
\label{finalForm}
\K_{\a}=\L_{\a}-\mathrm{tr}(\C_{\a})\C_{\a}-2\C_{\a}^2 =(\mu_4-3\mu_2^2)\Bigg(\begin{array}{cc} a_1^4+\frac{1}{2}(a_2^4+a_4^4) & \frac{1}{2}(a_2^4-a_4^4) \\  \frac{1}{2}(a_2^4-a_4^4) & a_3^4+\frac{1}{2}(a_2^4+a_4^4)\end{array}\Bigg).
\end{equation}
Finally, from \eqref{finalForm}, we get $\norm{\K_{\a}}^2=\sum_{k=1}^4  a_k^8 + (a^4_1+a^4_3)(a^4_2+a^4_4)$.

We note that the negative factor $(\mu_4-3\mu_2^2)$ in \eqref{finalForm} is in fact the kurtosis of $\rect(x)$, the sub-Gaussian constituent of the box spline. The fact that $\K_{\a}$ is negative-definite is thus consistent with the sub-Gaussian nature of the resulting box spline.

\subsection{Fast ZP interpolation}
\label{fast_zp}

\begin{figure*}
\centering
\includegraphics[width=0.6\linewidth]{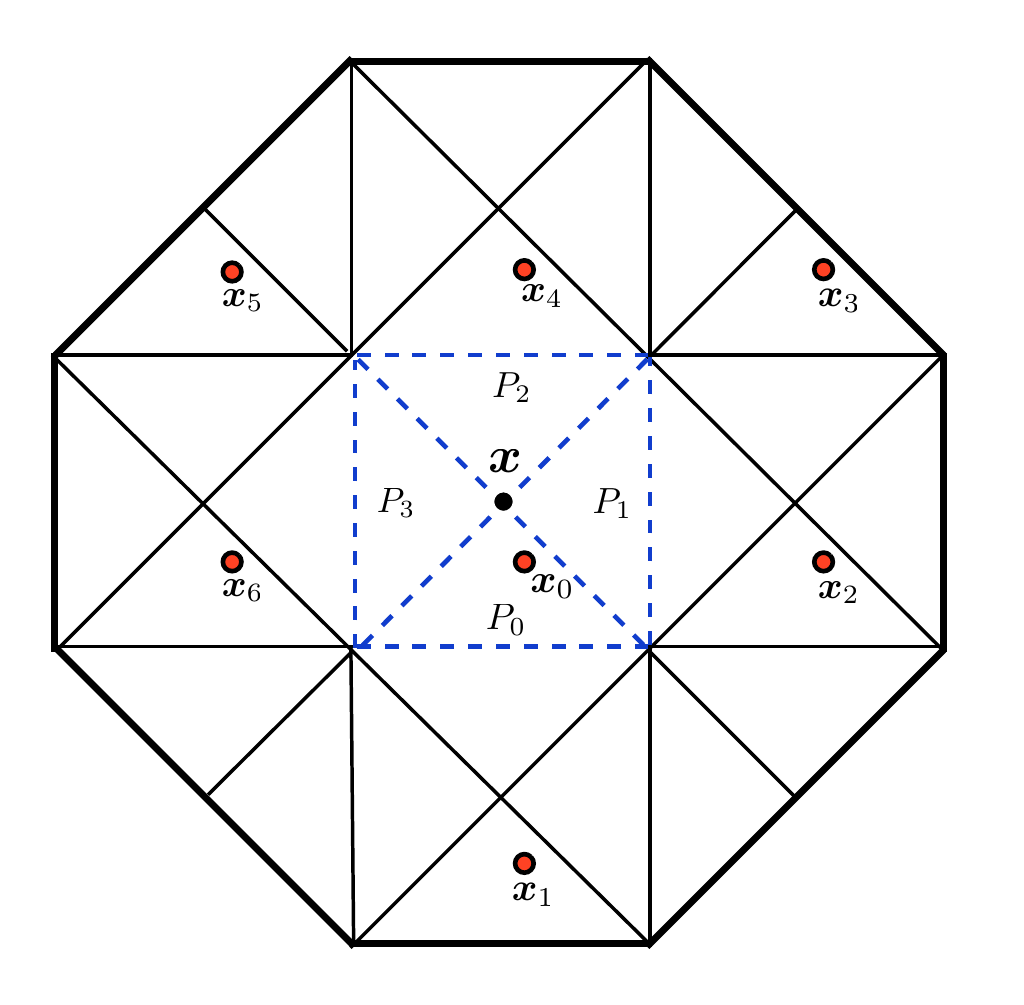}
\caption{The figure shows the translated ZP box spline $\bbeta^4_{\b}(\cdot+\x), \b=(1,\surd  2,1 ,\surd 2)$. The red dots $\x_1,\ldots,\x_7$ correspond to the points on the Cartesian lattice, and the triangular regions $P_1,\ldots,P_4$ are different partitions of the ZP, which together constitute a unit cell of the lattice.}
\label{interpolation}
\end{figure*}

	Given a discrete function $c[\n]$ and a point $\x$ on $\mathbf{R}^2$, we outline a technique for the fast evaluation of  the sum
\begin{equation}
\label{interpolation_sum}
\sum _{\n \in \mathbf{Z}^2} \c[\n] \bbeta^4_{\b}(\n-\x)
\end{equation}
where $\b=(1,\surd  2,1 ,\surd 2)$. A sketch of the partitions of the piecewise polynomial $\bbeta^4_{\b}(\x)$ is provided in Fig. \ref{interpolation}. The exact functional forms of the ZP box spline $2\bbeta^4_{\b}(\x)$ corresponding to these partitions can be found in \cite{Zwart}.  Since $\bbeta^4_{\b}(\x)$ has a compact support, this is in fact a finite sum, and requires at most seven evaluations of the  function $\bbeta^4_{\b}(\cdot-\x)$ for any arbitrary translation $\x$. This is illustrated in Fig. \ref{interpolation}, where the red dots $\x_0,\ldots,\x_6$ denote the lattice points that intersect the support of $\bbeta^4_{\b}(\cdot-\x)$. Thus, one needs to evaluate the translated ZP at the points $\x_0,\ldots,\x_6$ in order to compute the sum. The drawback here is that naive evaluation of the spline at every $\x_j$ requires a decision-making to figure out the associated partition before computing the corresponding polynomial. 

	The redundancy that we exploit is as follows: Consider the triangular regions $P_0,\ldots,P_3$ marked with blue dashed lines in Fig. \ref{interpolation} corresponding to the four different partitions of the ZP. These together constitute a unit cell of the lattice, and hence only one lattice point intersects them. The figure shows a particular instance where this point, denoted by $\x_0$, lies in $P_0$. This clearly fixes the partitions of the remaining lattice points $\x_1,\ldots,\x_6$. Thus, if we denote the polynomials corresponding to these partitions by $\rho_{0,0}(\x),\ldots,\rho_{0,6}(\x)$, then the sum in \eqref{interpolation_sum} is simply given by $\sum_{j=0}^6 c[\x_j]  \rho_{0,j}(\x_j)$. More generally, if $\x_0$ intersects the partition $P_i \ (0 \leq i \leq 3)$, and if we denote the corresponding polynomials by $\rho_{i,j}(\x)$, then the sum is given by $\sum_{j=0}^6 c[\x_j]  \rho_{i,j}\big(\x_j\big)$. Thus, we have the computational advantage that at most two binary decisions are required to simultaneously determine the ZP partitions corresponding to the points $\x_j$, where the coefficients of the polynomials $\rho_{i,j}(\x)$ can be pre-computed.

\section*{Acknowledgements}

	This work was supported by the Swiss National Science Foundation under grant 200020-109415, and the Ram\'on y Cajal program of the Spanish Ministry of Science and Innovation.

\bibliographystyle{amsplain}
\bibliography{refs_adaptive.bib}

\providecommand{\bysame}{\leavevmode\hbox to3em{\hrulefill}\thinspace}
\providecommand{\MR}{\relax\ifhmode\unskip\space\fi MR }
\providecommand{\MRhref}[2]{%
  \href{http://www.ams.org/mathscinet-getitem?mr=#1}{#2}
}
\providecommand{\href}[2]{#2}
\begin{thebibliography}{10}

\bibitem{Asahi}
T.~Asahi, K.~Ichige, and R.~Ishii, \emph{An efficient algorithm for
  decomposition and reconstruction of images by box splines}, {IEICE} Trans. on
  Fundamentals of Electronics, Commun. and Comp. Sc. \textbf{{E} 84} (2001),
  no.~8, 1883--1891.

\bibitem{deBoor}
C.~de Boor, K.~H\"ollig, and S.~Riemenschneider, \emph{Box {S}plines},
  Springer-Verlag, 1993.

\bibitem{ISBI_kunal}
K.~N. Chaudhury, Zs. P\"usp\"oki, A.~Mu{\~{n}}oz~Barrutia, D.~Sage, and
  M.~Unser, \emph{Fast detection of cells using a continuously scalable
  {M}exican-hat-like template}, {IEEE} International Symposium on Biomedical
  Imaging: From Nano to Macro (2010), in press.

\bibitem{Condat}
L.~Condat and D.~Van De~Ville, \emph{Three-directional box-splines:
  {C}haracterization and efficient evaluation}, {IEEE} Signal Process. Lett.
  \textbf{13} (2006), no.~7, 417--420.

\bibitem{Deriche}
R.~Deriche, \emph{Fast algorithms for low-level vision}, {IEEE} Trans. Pattern
  Anal. Mach. Intell. \textbf{12} (1990), no.~1, 78--87.

\bibitem{Entezari2008}
A.~Entezari, D.~Van De~Ville, and T.~M{\"o}ller, \emph{Practical box splines
  for reconstruction on the body centered cubic lattice}, IEEE Trans. Vis.
  Comput. Graph. \textbf{14} (2008), no.~2, 313--328.

\bibitem{Smeulders}
J.~M. Geusebroek, A.~W.~M. Smeulders, and J.~van~de Weijer, \emph{Fast
  anisotropic {G}aussian filtering}, {IEEE} Trans. Image Process. \textbf{12}
  (2003), no.~8, 938--943.

\bibitem{Heckbert}
P.~S. Heckbert, \emph{Filtering by repeated integration}, Intl. Conf. on
  Computer Graphics and Interactive Techniques \textbf{20} (1986), no.~4,
  315--321.

\bibitem{structure_tensor}
B.~J\"ahne, \emph{Digital image processing}, Springer, 1997.

\bibitem{Lam2007}
S.~Y.~M. Lam and B.~E. Shi, \emph{Recursive anisotropic 2-{D} {G}aussian
  filtering based on a triple-axis decomposition}, {IEEE} Trans. Image Process.
  \textbf{16} (2007), no.~7, 1925--1930.

\bibitem{Lee1980}
J.-S. Lee, \emph{Digital image enhancement and noise filtering by use of local
  statistics}, {IEEE} Trans. Pattern Anal. Mach. Intell. (1980), 165--168.

\bibitem{Arrate_TIP}
A.~Mu{\~{n}}oz-Barrutia, X.~Artaechevarria, and C.~Ortiz-de Solorzano,
  \emph{Spatially variant convolution with scaled {B}-splines}, {IEEE} Trans.
  Image Process. \textbf{19} (2010), 11--24.

\bibitem{Munoz}
A.~Mu{\~{n}}oz-Barrutia, R.~Ertl{\'{e}}, and M.~Unser, \emph{Continuous wavelet
  transform with arbitrary scales and {$\mathcal{O}(N)$} complexity}, Signal
  Processing \textbf{82} (2002), no.~5, 749--757.

\bibitem{PeronaMalik}
P.~Perona and J.~Malik, \emph{Scale-space and edge detection using anisotropic
  diffusion}, {IEEE} Trans. Pattern Anal. Mach. Intell. \textbf{12} (1990),
  no.~7, 629--639.

\bibitem{Richter}
M.~Richter, \emph{Use of box splines in computer tomography}, Computing
  \textbf{61} (1998), no.~2, 133--150.

\bibitem{Tan2003}
S.~Tan, J.~L. Dale, and A.~Johnston, \emph{Performance of three recursive
  algorithms for fast space-variant {G}aussian filtering}, Real-Time Imaging
  \textbf{9} (2003), no.~3, 215--228.

\bibitem{kurtosis}
A.~Tkacenko and P.P. Vaidyanathan, \emph{Generalized kurtosis and applications
  in blind equalization of {MIMO} channels}, Conference Record of the
  Thirty-Fifth Asilomar Conference on Signals, Systems and Computers \textbf{1}
  (2001), 742--746.

\bibitem{Wang1992}
X.~Wang, \emph{On the gradient inverse weighted filters}, {IEEE} Trans. Signal
  Process. \textbf{40} (1992), no.~2, 482--484.

\bibitem{Weikert1996}
J.~Weickert, \emph{Theoretical foundations of anisotropic diffusion in image
  processing}, Theoretical foundations of Computer Vision. Computing
  \textbf{11} (1994), 221--236.

\bibitem{Young}
I.~T. Young and L.~J. van Vliet, \emph{Recursive implementation of the
  {G}aussian filter}, Signal Processing \textbf{44} (1995), no.~2, 139--151.

\bibitem{Zwart}
P.~B. Zwart, \emph{Multivariate splines with nondegenerate partitions}, SIAM
  Journal on Numerical Analysis \textbf{10} (1973), no.~4, 665--673.

\end{thebibliography}

\end{document}